\documentclass[conference]{IEEEtran}

\usepackage{tikz}
\usepackage{amsmath}
\usepackage{amsfonts}
\usepackage{mathtools}
\usepackage{amsthm}
\usepackage{bm}
\usepackage{bbm} 
\usepackage[english]{babel} 
\usepackage{tcolorbox}

\newcommand{\dtoprule}{\specialrule{1pt}{0pt}{0.4pt}%
            \specialrule{0.3pt}{0pt}{\belowrulesep}%
            }
\newcommand{\dbottomrule}{\specialrule{0.3pt}{0pt}{0.4pt}%
            \specialrule{1pt}{0pt}{\belowrulesep}%
            }

\def\eqref#1{equation~\ref{#1}}

\def\1{\bm{1}}

\DeclareMathAlphabet{\mathsfit}{\encodingdefault}{\sfdefault}{m}{sl}
\SetMathAlphabet{\mathsfit}{bold}{\encodingdefault}{\sfdefault}{bx}{n}

\DeclareMathOperator*{\argmax}{arg\,max}
\DeclareMathOperator*{\argmin}{arg\,min}

\newcommand{\norm}[1]{\left\lVert#1\right\rVert} 

\theoremstyle{plain}
\newtheorem{theorem}{Theorem}[section]

\newtheorem{definition}[theorem]{Definition}

\newcommand{\circlednumber}[1]{%
\tikz[baseline=(char.base)]{
            \node[shape=circle,draw,inner sep=0.2pt] (char) {#1};}} 

\usetikzlibrary{patterns}

\pgfdeclarepatternformonly{dense north west lines}{\pgfqpoint{-1pt}{-1pt}}{\pgfqpoint{2pt}{2pt}}{\pgfqpoint{1.5pt}{1.5pt}}%
{
  \pgfsetlinewidth{0.4pt}
  \pgfpathmoveto{\pgfqpoint{0pt}{0pt}}
  \pgfpathlineto{\pgfqpoint{1.2pt}{1.2pt}}
  \pgfusepath{draw}
}

\newtcolorbox{kkbox}[1]{left=0.25mm, right=0.25mm, top=0.25mm, bottom=0.25mm, colframe=blue!66!black, boxrule=0.5pt, title={#1}, fonttitle=\bfseries, coltitle=blue!66!black, attach title to upper={\ }}

\usepackage{epsfig}
\usepackage{graphicx}
\usepackage{comment}

\usepackage{filecontents}

\usepackage{microtype}
\usepackage{graphicx}
\usepackage{caption}
\usepackage{subcaption}
\usepackage{booktabs}
\usepackage{blindtext}
\usepackage{threeparttable}
\usepackage{enumitem}
\usepackage{multirow}
\usepackage{multicol}
\usepackage[hidelinks]{hyperref}

\usepackage[capitalize,noabbrev]{cleveref}

\usepackage[textsize=tiny]{todonotes}

\newcommand{\revise}[1]{#1}

\newcommand{\ndssrevise}[1]{#1}

\usepackage{multirow}
\usepackage{multicol}
\usepackage{nicefrac}
\usepackage{microtype}
\usepackage{lipsum}
\usepackage{graphicx}
\usepackage{float}
\usepackage{algorithm}
\usepackage[noend]{algpseudocode}
\usepackage{booktabs}
\usepackage{makecell}
\usepackage{colortbl}
\usepackage{contour}
\usepackage{xspace}

\usepackage[normalem]{ulem}
\usepackage{soul}

\definecolor{Gray}{gray}{0.92}
\newcolumntype{g}{>{\columncolor{Gray}}c}
\newcolumntype{G}{>{\columncolor{Gray}}r}

\begin{document}

\title{CENSOR: Defense Against Gradient Inversion via Orthogonal Subspace Bayesian Sampling}

\hyphenation{op-tical net-works semi-conduc-tor}

\author{
\IEEEauthorblockN{
Kaiyuan Zhang, Siyuan Cheng,  Guangyu Shen, Bruno Ribeiro, 
\\
Shengwei An, Pin-Yu Chen$^{\dagger}$, Xiangyu Zhang, Ninghui Li}
\IEEEauthorblockA{Purdue University, $^{\dagger}$IBM Research}
\IEEEauthorblockA{\small \{zhan4057,\, cheng535,\, shen447,\, ribeirob,\, an93,\, xyzhang,\, ninghui\}@cs.purdue.edu,
\small {$^{\dagger}$}pin-yu.chen@ibm.com
}
}

\IEEEoverridecommandlockouts
\makeatletter\def\@IEEEpubidpullup{6.5\baselineskip}\makeatother
\IEEEpubid{\parbox{\columnwidth}{
    Network and Distributed System Security (NDSS) Symposium 2025\\
    24-28 February 2025, San Diego, CA, USA\\
    ISBN 979-8-9894372-8-3\\
    https://dx.doi.org/10.14722/ndss.2025.230915\\
    www.ndss-symposium.org
}
\hspace{\columnsep}\makebox[\columnwidth]{}}

\maketitle

\newcommand{\Tech}{{\textsc{Censor}\xspace}}

\begin{abstract}
Federated learning collaboratively trains a neural network on a global server, where each local client receives the current global model weights and sends back parameter updates (gradients) based on its local private data.
The process of sending these model updates may leak client's private data information.
Existing gradient inversion attacks can exploit this vulnerability to recover private training instances from a client's gradient vectors. Recently, researchers have proposed advanced gradient inversion techniques that existing defenses struggle to handle effectively.
In this work, we present a novel defense tailored for large neural network models. Our defense capitalizes on the high dimensionality of the model parameters to perturb gradients within a \textit{subspace orthogonal} to the original gradient. By leveraging cold posteriors over orthogonal subspaces, our defense implements a refined gradient update mechanism. This enables the selection of an optimal gradient that not only safeguards against gradient inversion attacks but also maintains model utility.
We conduct comprehensive experiments across three different datasets and evaluate our defense against various state-of-the-art attacks and defenses.
Code is available at \url{https://censor-gradient.github.io}.
\end{abstract}

\section{Introduction} \label{sec:introduction}

Federated learning (FL) gains its popularity as a privacy-preserving framework with many applications,
such as next word prediction~\cite{fedavg}, credit prediction~\cite{cheng2021secureboost}, and IoT device aggregation~\cite{samarakoon2018federated}, etc.
In FL~\cite{fedavg}, a global server broadcasts a global model to selected clients and collects model updates without directly accessing raw data.
On the client side, the model is locally optimized with decentralized private training data. 
Once the model updates are transmitted back to the server, an updated global model is constructed by aggregating individual received models. During the whole iterative training process, raw data will not be exchanged.

\begin{figure}[t]
    \centering
    \includegraphics[width=0.95\linewidth]{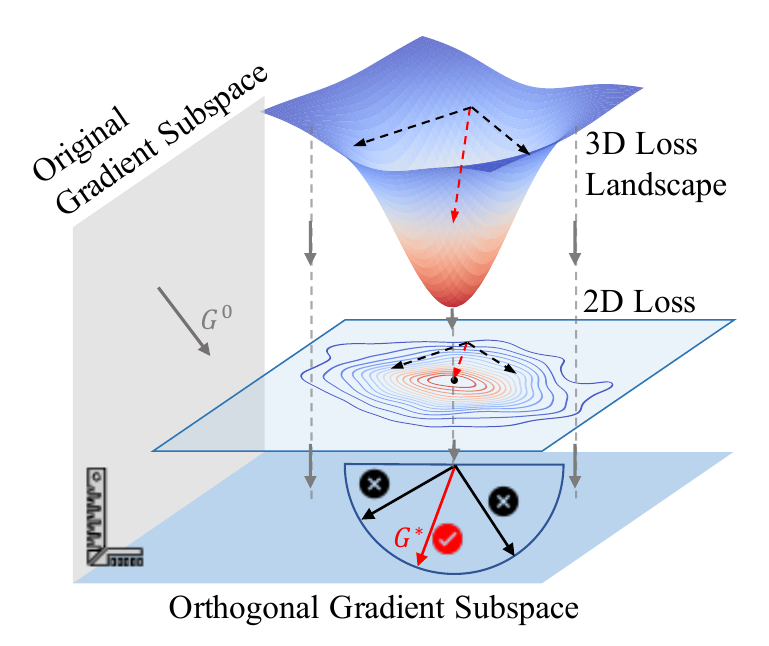}
    \caption{Intuition of cold Bayesian posteriors sampling over orthogonal subspace. The left \textcolor{gray}{gray plane} denotes the subspace that resides original gradients $G^0$.
    The bottom \textcolor{cyan}{blue plane} signifies a gradient subspace orthogonal to the original gradient subspace (\textcolor{gray}{gray plane}), denoted as $G^*$. In the middle of the figure, a 3D training loss landscape is projected onto the 2D plane, parallel to the orthogonal gradient subspace. Cold Bayesian posteriors sampling enables \Tech{} to sample multiple gradients and select the optimal one. This is illustrated by three potential directions in the orthogonal subspace, with the \textcolor{red}{red} indicating the optimal gradient that strictly points towards the optimal loss reduction direction.
    Notably, the optimal gradient highlighted in \textcolor{red}{red} not only lies within the orthogonal subspace, effectively protecting data privacy, but also minimizes the training loss, thereby ensuring the model utility.
    }
    \label{fig:intuition}
\end{figure}

Even though clients do not send their private training data in FL, the gradient updates they send are computed based on these private training data and can leak information in them.  
Recent studies have shown that private client data can be reconstructed through {\em  gradient inversion}~\cite{gifd, ig_geiping, gias, cafe_vfl, ggl, gi_yin, idlg, dlg}.
Inversion of image from gradient information was first discussed in~\cite{first_reconstruction}, which proved that recovery is possible from a single neuron or a linear layer.
In~\cite{dlg}, Zhu et al.~showed that accurate pixel-level inversion is practical for a maximum batch size of eight.
The attacker can either be an honest-but-curious server or an adversary who eavesdrops the communication channel between the server and the client to invert the private data of clients.
Our paper seeks to assess the attack risks and proposes methods to mitigate privacy leakage.

To mitigate the gradient inversion, researchers have proposed several defenses, including perturbing gradients with noise~\cite{dp_fl}, gradient clipping~\cite{clipping}, compressing gradients~\cite{compression}, and using specialized gradient representations~\cite{soteria}.
Although these techniques have demonstrated some degree of effectiveness against existing inversion attacks, it is likely that they will fail against attackers who either exploit more effective inversion attacks~\cite{gifd,gias,ggl} or have additional prior knowledge about the private data~\cite{ig_geiping,gi_yin}.  For example, if the attacker already suspects that the private instance is from a dataset known to the attacker, he can compute the gradient updates for each instance in the set and then identify which one matches the observed gradient.

In this paper, we propose a novel defense technique, \Tech{} (\textbf{\underline{C}}old post\textbf{\underline{E}}riors co\textbf{\underline{N}}trolled \textbf{\underline{S}}ampling over \textbf{\underline{OR}}thogonal subspace), 
to address the dual challenge of preserving privacy while maintaining the utility of machine learning models through a novel gradient refinement methodology. 
We refine model parameters by selectively updating them along an \textit{orthogonal subspace}~\cite{orth_subspace,subspace_learning} with \textit{cold Bayesian posteriors}~\cite{izmailov2021bayesian, neal2012bayesian}.
The key insight is that Bayesian posterior sampling, which does not rely on gradient-based optimization, better resists gradient-based inversion attacks of a client's private data.
And sampling inside an orthogonal subspace to the model gradients over the private data, ensures that the resulted gradients do not resemble the original sensitive ones.

\textit{Bayesian posterior sampling} aims to sample the model parameters from the posterior distribution $P(\theta|D)$, which represents the probability of the parameters $\theta$ given the data $D$. This allows a client to obtain model parameters without resorting to gradients.
However, sampling from $P(\theta|D)$ can lead to parameter samples that fit the data poorly.
Cold Bayesian posteriors modify the distribution $P(\theta|D)$ by introducing a \textit{temperature} parameter $M$. 
By tuning $M$, typically $0 < M \ll 1$, where $M \approx 0$ is the coldest temperature, the distribution becomes sharply peaked around the parameters that significantly reduce the loss, akin to concentrating the distribution around a Maximum a Posteriori (MAP) estimate, but retaining the ability to sample around the MAP solution.

An \textit{orthogonal subspace} is a vector space where each vector is perpendicular to the vectors of another subspace.
In our scenario, we define orthogonal subspace as the one that is perpendicular to the client's original gradient subspace.
Restricting our parameter sampling to an orthogonal subspace means we will not inadvertently return the original gradients (unlike some methods as described in~\cite{dp_fl}).
More precisely, in a model with $m$ neuron weights, this orthogonal subspace can be defined by a set of $m-k$ linearly independent vectors orthogonal to the gradients of $k$ training data instances.
And, since the orthogonal subspace of a gradient of a model with $m$ parameters has $(m-k)$-dimensions, and the resulting output of a client is a single vector in this subspace (a single gradient), the potential leakage of the original gradients is negligible if $m \gg k$.

By leveraging \textit{cold posteriors over orthogonal subspaces}, \Tech{} employs a refined gradient update mechanism. By evaluating the impact of a set of gradients inside the orthogonal subspace of the instance gradients and selecting those that sufficiently reduce the loss, our method mitigates the privacy leakage and in the mean time, keeps improving the utility of the model.
Figure~\ref{fig:intuition} provides an intuitive illustration of our technique.

We summarize our contributions as follows:
\begin{itemize}

    \item We introduce \Tech{} (\textbf{\underline{C}}old post\textbf{\underline{E}}riors co\textbf{\underline{N}}trolled \textbf{\underline{S}}ampling over \textbf{\underline{OR}}thogonal subspace), a defense mechanism against gradient inversion attacks that operates by posterior-sampling model updates from a subspace orthogonal to the original gradients subspace. This technique does not rely on gradients and, therefore, better resists gradient reconstruction attacks, thereby preserving the privacy of the model.

    \item We enhance the balance between utility and privacy by incorporating \textit{cold Bayesian posteriors} into our methodology. By adjusting the {\em temperature} parameter, we refine the selection process for gradient updates, narrowing the focus to those gradients that optimally maintain utility while minimizing privacy risks.

    \item We conduct extensive experiments to evaluate the effectiveness of \Tech{} against state-of-the-art gradient inversion attacks. Our empirical results demonstrate that \Tech{} not only mitigates potential privacy breaches more effectively but also outperforms state-of-the-art defenses across multiple quantitative and qualitative metrics.

\end{itemize}

\smallskip
\noindent
\textbf{Threat Model} \label{sec:threat_model}
\ndssrevise{
Our threat model is consistent with the literature~\cite{ig_geiping, gi_yin, ggl, gias, gifd, soteria, compression}, where the adversary is considered as an honest-but-curious server. 
This adversary aims to invert training samples without direct access to the data from local clients or original training data.
}
The adversary knows the model architecture and local gradients transmitted by clients, while the adversary can not modify either the model or the gradients~\cite{fowl2021robbing, wen2022fishing}. 
Additionally, we assume the adversary can utilize the knowledge extracted from publicly available datasets and leverage pre-trained neural network models, such as those based on Generative Adversarial Networks (GANs) to facilitate the attack.
The benign local clients can conduct defense to protect their data privacy. They have access to the model parameters at each training round but have no knowledge about the adversary's attack configuration or technique.
In the end, our analysis assumes the most favorable conditions for the adversary, setting the batch size as one~\cite{ggl,gifd}.
We also adopt loose restrictions regarding the adversary, assuming they possess sufficient computational power and memory.

\smallskip
\noindent
\textbf{Roadmap.} The rest of this paper is organized as follows. 
In Section~\ref{sec:preliminaries}, we formulate the problem and introduce the background of gradient inversion attacks and defenses.
In Section~\ref{sec:observations}, we discuss two new observations about existing gradient inversion attacks.
In Section~\ref{sec:method}, we present the theoretical
analysis for our technique and introduce the detailed design of our defense.
In Section~\ref{sec:evaluation}, we present a comprehensive experimental evaluation of \Tech{} against various attacks and compare our proposed defense with state-of-the-art defense baselines.
In Section~\ref{sec:related_work}, we review related literature.
In Section~\ref{sec:conclusion}, we offer concluding remarks.
We also provide a summary of all notations in Appendix~\ref{sec:appendix}, Table~\ref{tab:notation} for easy reference.

\section{Background} \label{sec:preliminaries}
In this section, we start by formulating the gradient inversion within the context of federated learning. Following this, we offer a brief overview of various existing attack techniques and several defense methods designed to counter them.
Additionally, we discuss the limitations of these existing defense strategies, underscoring the necessity of introducing our approach.

\subsection{Problem Formulation}
We focus on the setting of learning a classifier using federated learning.
Given a neural network $f_{\theta}$ parameterized by $\theta$, the training objective is to obtain model parameters through empirical risk minimization
\begin{equation}\label{equ:overall}
    \hat{\theta} = \argmin_{\theta} \sum_{j} \ell (f_{\theta}(X_j), Y_j),
\end{equation}
where $\ell$ is the loss function, and $X_j$, $Y_j$ are the input features and the corresponding label of the $j$-th training instance, respectively.

Within a federated learning framework, a \textit{global server} aims to solve \Cref{equ:overall} through collaborative training with multiple \textit{local clients}, each possessing a subset of the training data.  Training consists of many iterations.  In the $\tau$-th iteration, the server sends the parameter $\theta_{\tau}$ to each client.  The $k$-th client then computes and sends the following gradient to the server
\begin{equation}
    G_{k}^\tau = \sum_{j} \nabla \ell (f_{\theta_{\tau}}(X_{j,k}), Y_{j,k}), 
\end{equation}
and the server updates its model through
\begin{equation}
    \theta_{\tau+1} = \theta_{\tau} - \eta \sum_{k=1}^{N} G_k^\tau ,
\end{equation}
where $N$ is the number of clients, and $\eta$ is the learning rate. Before sending the gradient $G_k^\tau$ back to the server, local clients have the opportunity to apply defense mechanisms to it.

\subsection{Gradient Inversion Attacks in Federated Learning.}
Gradient inversion attacks present a significant threat to data privacy in federated learning systems. It enables adversaries to reconstruct private data samples of clients with high fidelity.
Existing attacks can be broadly categorized into two groups: (1) Stochastic Optimization Attacks, and (2) GAN-based Attacks.

\smallskip \noindent
\textbf{Stochastic Optimization Attacks.}
Attacks based on stochastic optimization invert training images from random initialization with the guidance of gradients and potential other prior knowledge.
A famous work~\cite{dlg} conceptualized the attack vector as an iterative optimization challenge, where attackers approximate original data samples by minimizing the discrepancy between actual shared gradients and synthetic gradients derived from artificially generated data samples.
Several following works have refined this attack methodology. For instance, \cite{idlg} introduced a technique to infer the labels of individual data samples directly from their gradients, providing more guidance towards the ground-truth.
Moreover, \cite{ig_geiping} achieved the inversion of higher-resolution images from sophisticated models like ResNet by modifying the distance metric used for optimization and incorporating a regularization term to the process.

Following existing works~\cite{ig_geiping, gi_yin}, we define the gradient inversion using stochastic optimization as follows:
Given a neural network with parameters $f_\theta$, the attacker initiates the process by generating a single instance random noise $X'$ and its label $Y'$ as its initialization. The instance is iteratively refined to align with the ground-truth local gradients $g$.
For simplicity, we define a function $F(\cdot)$ as deriving the gradients of the model $f_\theta$ with regards to the input, where
\begin{equation}
    F(X') = \nabla\ell(f_{\theta}(X'), Y').
\end{equation}
There are several methods (which will be discussed later) for inferring the label $Y'$ before the optimization process, so we exclude it from $F(\cdot)$.
The stochastic optimization is driven by the goal of minimizing the discrepancy between the dummy gradients $F(X')$ and the original local gradients $g$ that were submitted by the benign client, as described below:
\begin{equation}\label{equ:obj_function}
    \hat{X'} = \argmin_{X'}   \mathcal{D} \left ( F(X'), g \right )
\end{equation}
where $\mathcal{D}(\cdot, \cdot)$ is the distance metric, e.g. $l_2$-distance~\cite{gi_yin}.

\smallskip \noindent
\textbf{GAN-based Attacks.}
Recent attacks~\cite{gifd, ggl, xia2022gan} leverage Generative Adversarial Networks (GAN)~\cite{goodfellow2014generative} to facilitate the reconstruction of high-quality images.
GAN is a deep generative model, which is able to learn the probability distribution of the images from the training set.
In~\cite{gias}, Jeon et al.~proposed to search the latent space and parameter space of a generative model, effectively harnessing the generative capabilities of GANs to produce high-quality inverted images. 
However, it requires a specific generator to be trained for each inverted image, which may consume large amounts of GPU memory and inference time.
A follow-up study~\cite{gradvit} extends attacks on Vision Transformers. 
In addition, \cite{ggl} adopted the generative model with label inference, which achieves semantic-level inversion. Among these GAN-based methods, only~\cite{gias} addresses the scenario where the training data for the generative and global models come from different probability distributions.
Intermediate Layer Optimization(ILO)~\cite{daras2021intermediate} also proposes an optimization algorithm for solving inverse problems with deep generative models. 
Instead of solely optimizing the initial latent vectors, they progressively alter the input layer, resulting in increasingly expressive generators.
Following ILO, Fang et.al~\cite{gifd} exploits pre-trained generative models as data prior to invert gradients by searching the latent space and the intermediate features of the generator successively with $l_1$ ball constraint, address the challenges of expression ability and generalizability of pre-trained GANs.

\revise{
Following existing works~\cite{ggl, gias, gifd}, we define GAN-based attacks by assuming that the attacker has access to a pre-trained generative model, which is trained using a large public dataset. The problem can be formulated as follows:
\begin{equation}
    z^* = \argmin_{z \in \mathbb{R}^{k}} \mathcal{D}(\mathcal{T}(F(G_p(z))), g) + \phi(G_p; z),
\end{equation}
where $G_p$ denotes the pre-trained generative model and $z \in \mathbb{R}^{k}$ represents its latent space.
$\mathcal{T}$ is a gradient transformation function such that the attackers can adaptively counter the effects of defense strategies.
$\phi(\cdot)$ represents a regularization term that imposes penalties on latent vectors that diverge from its original prior distribution of $G_p$.
}

\revise{
\smallskip \noindent
\textbf{Label Inference Assisting Gradient Inversion.}
Several existing attacks, both stochastic and GAN-based, leverage label inference to enhance the fidelity of inversion.
These attacks show that private labels can be directly inferred from the gradients~\cite{idlg, gi_yin, ggl}.
Utilizing the shared gradients, the adversary initially employ analytical inference techniques~\cite{idlg, ggl} to infer the ground-truth label $c$ of the client’s private image. 
Adopting notation from~\cite{ggl}, we present the label inference process for federated learning (FL) models engaged in a classification task across $C$ classes.
The computation for the $i$-th entry of the gradients associated with the weights of the final fully-connected (FC) classification layer (denoted as $\nabla W_{\text{FC}}^{i}$) is described by the following equation:
\begin{equation}
    \nabla W_{\text{FC}}^{i} = 
    \frac{ \partial  \ell (f_{\theta^{\tau}}(X), Y)}{ \partial z_i} \times \frac{ \partial z_i}{ \partial  W_{\text{FC}}^{i}}
\end{equation}
where $z_i$ denotes the $i$-th output of the FC layer.
Note that the calculation of the second term, $\frac{ \partial z_i}{ \partial  W_{\text{FC}}^{i}}$, yields the post-activation outputs from the preceding layer.  
When activation functions such as ReLU or sigmoid are employed, these outputs will always be non-negative.
When networks are trained with cross-entropy loss using one-hot labels (and assuming softmax activation is applied in the final layer), the first term will only be negative when $i = c$.
Consequently, the ground-truth label can be identified by locating the index where $\nabla W_{\text{FC}}^{i}$ is negative.
}

\subsection{Defenses Against Gradient Inversion Attacks}
Next, we introduce several current defense mechanisms against gradient inversion attacks.

\smallskip \noindent
\textbf{Noisy Gradient.}
Differential privacy (DP) is the established method for quantifying and controlling the privacy exposure of individual clients. 
In federated learning, DP~\cite{dp_fl} can be applied at either the server’s side or the client’s side. 
Clients can employ a randomized mechanism to modify the gradients before they are shared with the server.
DP provides a theoretical worst-case guarantee about the amount of information an adversary can extract from the data released.
However, achieving these worst-case guarantees often requires substantial additive Gaussian noise to the gradient~\cite{wei2021user, abadi2016deep}, which may significantly compromise the utility of the global model.

\begin{definition} [Differential Privacy]
A randomized mechanism $\mathcal{M} : \mathcal{A} \rightarrow \mathcal{R}$ with domain $\mathcal{A}$ and range $\mathcal{R}$ satisfies $(\epsilon, \delta)$-differential privacy if, for any two adjacent inputs $d, d' \in \mathcal{A}$ and for any subset of outputs $S \subseteq \mathcal{R}$, it holds that
\[
\Pr[\mathcal{M}(d) \in S] \leq e^\epsilon \Pr[\mathcal{M}(d') \in S] + \delta,
\]
where $\epsilon > 0$ is the privacy budget and $\delta \geq 0$ is the privacy loss parameter.
\end{definition}

\smallskip \noindent
\textbf{Gradient Clipping.}
Wei et al.~\cite{clipping} introduces a defense mechanism resilient to gradient leakage, utilizing a client-level differential privacy (DP) approach.
In each layer $l$, the gradient $g_l$ is computed for every individual training example, and the clipping transformation function is defined as
\begin{equation}
    \mathcal{T}(g_l, b) = g_l \cdot min(1, b/\norm{g_l}_2),
\end{equation}
where $b$ is the constant representing the upper bound for clipping.
This method applies per-example gradient clipping, followed by the addition of a Gaussian noise vector to the clipped gradient, to secure demonstrable differential privacy guarantees. 
Wei et al.~have demonstrated that gradient clipping can achieve client-level per-example DP and that the approach is resistant to gradient inversion.
However, existing attacks~\cite{ggl, gifd} have empirically indicated that gradient clipping defense mechanisms are often ineffective in federated learning contexts.
The reason is that the attacker can estimate the clipping bound by computing the $l_2$ norm of the gradients received for each layer, thereby revealing strong guidance towards the ground truth gradient.

\smallskip \noindent
\textbf{Gradient Compression / Sparsification.}
In Top-$k$ compression~\cite{compression}, the client only selects the $k$ largest components of a gradient $g$ in terms of absolute values and sets all other entries to zero.
This technique defines the compressed gradient $g'$ such that each component $g'_i$ of $g'$ is determined by the following:
\begin{equation}
    g'_i = 
    \begin{cases} 
        g_i & \text{if } g_i \in \text{Top}_k(\{|g_1|, |g_2|, \ldots, |g_n|\}) \\
        0 & \text{otherwise}
    \end{cases}
\end{equation}
This operation effectively retains only the $k$ components with the highest magnitudes for gradient transmission.
Existing attacks~\cite{dlg, ggl, gifd} have shown that gradient sparsification is not always effective.
Since the implementation involves applying a gradient mask, the attackers is able to estimate the gradient's sparsity by observing the percentage of non-zero entries in the shared gradients, thereby leaking information of the ground-truth gradients.

\smallskip \noindent
\textbf{Gradient Representation.}
Recent work~\cite{soteria} identifies gradient-induced data leakage in federated learning and introduces a defensive strategy called Soteria.
This approach calculates gradients using perturbed data representations.
Let $X$ and $X'$ represent the original and inverted images via perturbed gradients, respectively, with $r$ and $r'$ as their corresponding representations in the protected layer. 
Soteria aims to minimize information leakage by maximizing the distance between $X$ and $X'$ while ensuring that the representations $r$ and $r'$ remain similar. 
The optimization problem is formulated with constraints as follows:
\begin{equation}
\begin{aligned}
& \max_{r'} \quad \left\| X - X' \right\|_2, \\
& \text{s.t.} \quad \left\| r - r' \right\|_0 \leq \epsilon
\end{aligned}
\end{equation}
In Soteria, the $l_0$ norm ensures sparsity, effectively acting as a pruning rate in the gradient sparsification mechanism. 
The defense involves an optimization phase that introduces substantial computational overhead, particularly within the fully-connected layers. 
Despite this, Soteria maintains robustness and does not hinder the convergence of federated learning systems~\cite{soteria}.
However, the defense mechanism is essentially equivalent to apply a gradient mask to the protected layer.
Once the global model $f_{\theta}$ and input $X$ are given, the masking process is deterministic. 
Consequently, an attacker could easily reverse engineer the original gradients in the protected layers.

\section{Observations on Existing Attacks} \label{sec:observations}
The literature observes several effective gradient inversion attacks that underscore significant privacy threats. At the same time, researchers have identified limitations in these attacks.
For example, Huang et al.~\cite{huang2021evaluating} identified two assumptions that IG~\cite{ig_geiping} and GI~\cite{gi_yin} rely upon for effectiveness.  First, these attacks assume that the adversary knows the batch normalization (BN) statistics, such as mean and variance of the input instance to be constructed.  Second, they assume the adversary knows or is able to precisely infer the label of input instance and use that information in reconstruction.
These information is typically unavailable in practical real-world scenarios.
It was shown that by nullifying these assumptions, the performance of the attacks degrades significantly, only working for low-resolution images.

In this section, we further introduce two additional observations, serving as the basis for our defense design.
We first implement and evaluate five state-of-the-art gradient inversion attacks, including stochastic optimization attacks IG~\cite{ig_geiping} and GI~\cite{gi_yin}, and GAN-based attacks GGL~\cite{ggl}, GIAS~\cite{gias}, and GIFD~\cite{gifd}.
We aim to better understand the strengths and limitations of these gradient inversion attacks, in order to help us more effectively defend against them.
We focus on the setting where the clients use a batch size of one; i.e., each client sends to the server the gradient of a single instance and the adversary tries to reconstruct this instance from the gradient.
We prioritize this setting because we aim to develop defense mechanisms capable of countering attacks in this particularly challenging context, which is the easiest case to attack and thus the hardest to defend against.

In our study, we make two new observations about existing inversion attacks:
\protect\circlednumber{1} Most existing attacks succeed only within the early stage of training;
\protect\circlednumber{2} GGL is the only attack that is able to produce \textit{high-quality} images beyond the first few epochs.
GGL achieves this by first inferring the label of the input instance; however, it typically reconstructs images that are independent of the ground-truth input instance.
Taken these two observations together, we conclude that the threat of gradient inversion attacks mostly lies during the first few training epochs and that the defense needs to be improved to protect the label leakage.
Next, we elaborate on these observations.

\begin{figure}[ht]
    \centering
    \includegraphics[width=.48\textwidth]{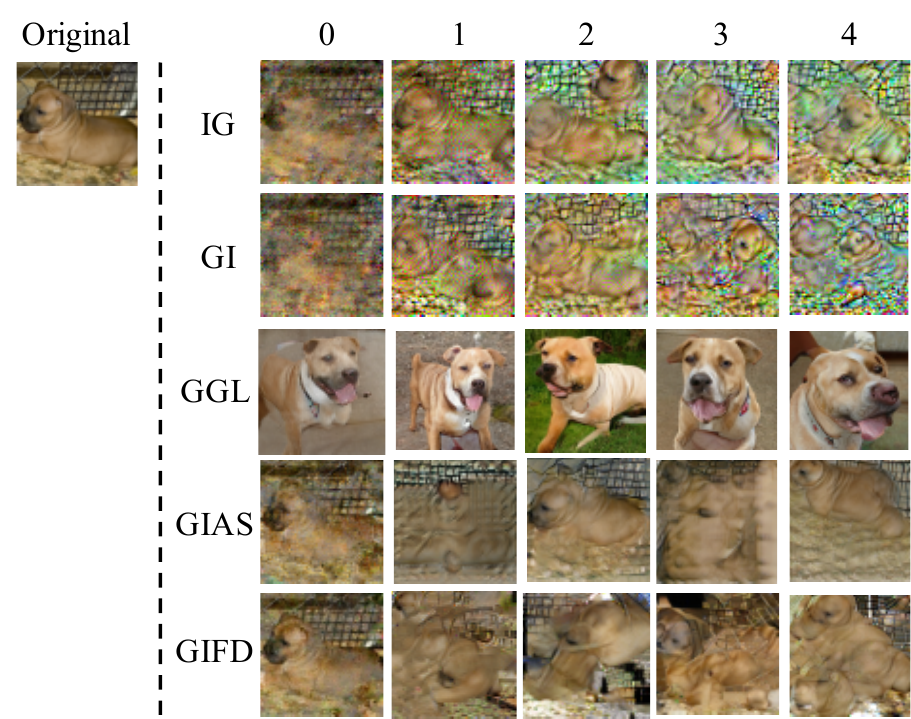}
    \caption{Results of SOTA attacks inversions (batch size equals to 1) across the initial five epochs.}
    \label{fig:obser_first_5ep}
\end{figure}

\smallskip \noindent
\textbf{\underline{Observation \protect\circlednumber{1}}: Most existing attacks succeed only within the early stage of training.}\label{sec:existing_fail_first_few}
Most existing attacks~\cite{ggl,gias,gifd} only evaluate the inversion performance at the first epoch 0, assuming random parameter initialization.
To evaluate the robustness of these attacks in a more realistic federated learning scenario, we design an experiment with over a thousand clients. 
Each client processes data with a batch size of 1. 
To speed up the training within this simulation, 
we employ a setup consisting of one \textit{victim} client with a batch size of 1, and another local client with a larger batch size of 1024.
Both clients utilize a stochastic gradient descent (SGD) optimizer, which is commonplace in standard federated learning frameworks~\cite{fedavg}.
As the number of training rounds increase, the ability of an attacker to invert and extract useful information diminishes.
This configuration suggests relative safety for benign clients when submitting gradients during the later converged stages of training. 
\emph{However, the inherently \textit{non-i.i.d.} nature of federated learning often results in scenarios where some clients possess only one or a few training instances.}
Concern arises when such a benign client participates in the initial stages of federated learning. 
Besides, batch size of 1 is the easiest setting to attack and the hardest to defend against. 
The security of this client’s raw data against inversion attacks, under these conditions, remains questionable. 
Given these complexities, it is imperative to evaluate the security implications more rigorously.

The results of the experiment is illustrated in Figure~\ref{fig:obser_first_5ep}, where each row corresponds to a different attack, with columns to the right of the dotted line representing the rounds of federated learning training.
Note that no defense is applied here, and we are investigating the vanilla performance of the various attacks.
We show the experiment result using a batch size of 1 over 5 FL training rounds.
Observe that while certain attacks successfully invert data during the initial epochs (e.g., the first two), others are confined to reconstructing only partial details (such as eyes or texture features). 
Notably, we find that the only attack that is able to produce high-quality images, but not faithfully same as the original instances in the first few epochs, is GGL~\cite{ggl}, as shown in the third-row image of Figure~\ref{fig:obser_first_5ep}.
This phenomenon is further explored in the subsequent observation.

This empirical evidence underscores a critical insight:
\begin{kkbox}{\small Privacy leakage persists, even though the effectiveness of attacks decreases in the early training epochs.}\label{obser:observation1}
{\em \small  \textit{Securing the initial training phase for benign local clients, especially at batch size of 1 for the easiest setting to attack, poses significant challenges.
}}
\end{kkbox}

In light of this, our defense strategy, \Tech{}, 
is strategically implemented within the first few epochs.
In our evaluations, we also opt for conditions most favorable to the attacker, specifically a batch size of one, to demonstrate the robustness of our approach in protecting against gradient inversion attacks.

\begin{figure}[ht]
    \centering
    \includegraphics[width=.48\textwidth]{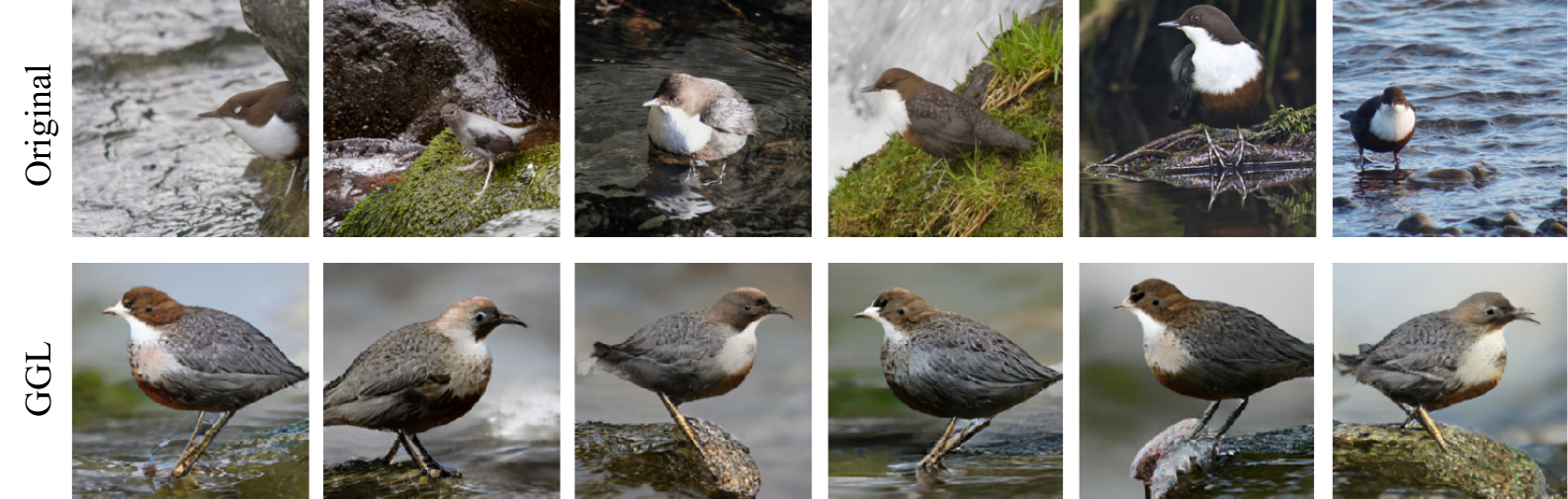}
    \caption{GGL consistently inverts similar images across distinct inputs.
    }
    \label{fig:moti_similar_inversion}
\end{figure}

\smallskip \noindent
\textbf{\underline{Observation \protect\circlednumber{2}}: GGL~\cite{ggl} inverts high-quality but typically low-fidelity images, and reveals label information.}
From the above analysis and the illustrations in Figure~\ref{fig:obser_first_5ep}, the GGL~\cite{ggl} attack uniquely manages to produce high-quality images even after the initial epochs.
While one could interpret this as GGL being an especially effective gradient inversion attack, from Figure~\ref{fig:obser_first_5ep} we observe that the reconstructed images, while being clear representatives of the label of the original instance, do not look particularly similar to the original instance.  

We conduct additional experiments to analyze this effect. Figure~\ref{fig:moti_similar_inversion} shows the result, where the first row shows six different training instances of the class bird, and the second row shows the corresponding instances reconstructed by GGL using gradients submitted in the first epoch 0.
Note that while the second row shows high-quality bird images, these images are similar to each other, and notably distinct from the ground-truth image in the first row.
This is because GGL is able to successfully infer the label of the original instance, and then use the robust generative capabilities of GANs to produce well-formed images that are good representatives of the label.
On the other hand, while GGL may not faithfully reconstruct the original training samples, it does identify the ground-truth label and provides representative instances of that label.
Therefore, it still to some extent compromises the privacy of the local training data.

This empirical evidence highlights a critical insight:
\begin{kkbox}{\small Privacy leakage persists, GGL inverts low-fidelity images, yet reveals label information.
}\label{obser:observation2}
{\em \small  \textit{
The GGL attack, despite primarily producing low-fidelity images, effectively infers label information and provides label-representative instances through its strategic exploitation of the latent space using a pre-trained GAN. This capability highlights a significant challenge for defense mechanisms, which struggle to adequately prevent label leakage despite existing efforts to obfuscate or alter the data representation.
} 
}
\end{kkbox}

Our proposed methodology, \Tech{}, introduces a novel defensive mechanism that strategically employs an alternative gradient vector located within a high-dimensional space, specifically within a subspace orthogonal to the original gradient vector.
In other words, \Tech{} presents a radically different gradient profile from an orthogonal subspace, which results in seriously corrupted gradient information, rendering GGL unable to precisely infer the label information and perform any effective inversion.

\smallskip \noindent
\textbf{\underline{Summary.}}
Figure~\ref{fig:obser_first_5ep} shows that all state-of-the-art attacks other than GGL reconstructs low-quality images except for the first few epochs, and Figure~\ref{fig:moti_similar_inversion} shows that GGL's high-quality reconstructed images are not high-fidelity reconstructions of the input instance, but rather representative instances of the label. Going forward, we thus focus on defending against gradient inversion attacks during the first few training epochs and enhance the protection against label leakage.

\section{Proposed Defense: Gradient Sampling over Orthogonal Subspaces} \label{sec:method}
In this section, we first lay foundation of the theoretical analysis for our defense technique, \Tech{}.
Then we introduce the details of our proposed defense, which takes advantage of the high dimensionality of the model parameters to sample gradients in a \textit{subspace} that is \textit{orthogonal} to the original gradient, such that the new gradient also reduces the loss like the original gradient but is not in the same direction.

\subsection{Learning over Orthogonal Subspaces}

\noindent
\textbf{Rationale: Learning through Bayesian Updates on Orthogonal Subspaces Instead of Gradients.}
Statistically, the minimization described in \Cref{equ:overall} that reduces the loss $\ell(\cdot)$ ---cross-entropy or L2 loss--- is equivalent to maximizing the log-likelihood of categorical or Gaussian distributions~\cite[pp.\ 25--27]{bishop2006pattern}, which is also equivalent to minimizing the negative log-likelihood of the (neural network) model.

Maximizing the likelihood in neural networks uses gradients to update the model.
Interestingly, in contrast to maximizing the
likelihood, which seeks to find a single optimal set of neuron weights, a Bayesian approach yields a posterior distribution $P(\theta|D)$ over the neural network parameters $\theta$, conditional on the training data $D$.
The neural network parameter prior $P(\theta)$ is usually a standard (isotropic) Gaussian.
This posterior distribution can be sampled and the resulting sampled parameters can be used for predictions.

The key insight of \Tech{} is to develop a Bayesian-inspired procedure to update the global model that is resistant to gradient inversion attacks. If client $k \in \{1,\ldots,K\}$ can somehow sample from a modified posterior $P(\theta'|D^k)$ (not necessarily the true posterior $P(\theta|D^k)$) to produce updates to the global model, the client can send the global model an update that is not based on gradients. {\em If the update is not a gradient, then gradient inversion attacks would struggle to invert it}.

In order to avoid gradients, we first need to overcome some obstacles:

{\em In the following exposition, the terms $P(D|\theta)$ and $P(\theta|D)$ are described as normalized probabilities. It is important to note, however, that the equations are equally applicable to probability densities and unnormalized probabilities.}

\smallskip \noindent
\textbf{\underline{\em Obstacle 1:}} Bayesian neural networks~\cite{mackay1992practical,neal2012bayesian} offer a gradient-free alternative to maximum likelihood, but they are not without drawbacks~\cite{zhang2019cyclical}.
In overparameterized neural networks, even if the posterior probability $P(\theta|D)$ of sampling poorly-performing neural network parameters is small ---i.e., $f_{\theta}$ has a large loss (equivalently a small probability $P(D|\theta')$)--- in overparameterized neural networks these are still likely to be sampled due to the high dimensionality of $\theta$. 

A simple solution is noticing that the maximum a-posteriori (MAP) solution $\hat{\theta}_p = \argmin_{\theta}  - \log P(\theta|D) = \argmin_{\theta} - \log P(D|\theta) - \log P(\theta)$ is very close to the maximum likelihood estimation (MLE) solution $\hat{\theta} = \argmin_{\theta}  - \log P(D|\theta)$ obtained by minimizing the loss with gradient descent, if the prior $P(\theta)$ is weak (which is our case)~\cite{neal2012bayesian}. 
However, employing MAP reverts the model update back to gradient-based optimization.
Thankfully, there is a middle-ground: {\em cold posteriors}. Cold posteriors are a hybrid between sampling parameters from the original posterior $P(\theta|D)$ and the MAP gradient-based solution. Cold posteriors  make use of the fact that the modified posterior $P^M(\theta|D) \propto P(D|\theta)^M P(\theta)$ is highly concentrated around the MAP solution $\hat{\theta}_p$ when the ``temperature'' parameter $M$ is close to zero but still allows Bayesian sampling~\cite{izmailov2021bayesian}.
By setting the temperature parameter $M$ close to zero, we can sample neural network parameters from the Bayesian posterior without sacrificing too much the quality of the model.

\smallskip \noindent
\textbf{\underline{\em Obstacle 2:} The global server expects gradients from its clients.}
Here, we design an acceptance/rejection sampling procedure inspired by Metropolis-Hastings's rejection sampling that outputs parameter updates that can be used like gradients by the global server. By sampling a perturbation $G \sim \mathcal{N}({\bf 0},\epsilon I)$ to the current global model parameters $\theta_\tau$, for $\epsilon \in R^+$ small, and then accepting the sample $G$ with probability 
\[
P(\text{accept}) = \min\left(1,\frac{P(D^k | \theta_\tau + G)^M P(\theta_\tau + G)}{P(D^k | \theta_\tau)^M P(\theta_\tau)}\right),
\]
where $\theta_\tau$ is the current global model at step $\tau$ at client $k$ and $M > 0$ is the temperature described earlier; otherwise, restart the sampling-acceptance process. The Metropolis-Hastings samples behave as if sampled from $P^M(\theta|D)$.
Using rejection sampling will allows us to have some flexibility defining the ``gradient'' perturbation $G$ later. 

Our approach avoids rejections by sampling $T$ proposed random perturbations $G'^1,\ldots,G'^T$ and, since we are interested in cold posteriors, client $k$ simply selects the best proposal $G^\star = \argmax_{G \in \{G'^1,\ldots,G'^T\}} P(D^k | \theta_\tau + G)$.
This approach, however, has a potential challenge: For $T \gg 1$, $G^\star$ may be similar to the MAP gradient at $\theta_\tau$, which could be inverted by the adversary. In what follows we address this potential drawback.

\smallskip \noindent
\textbf{\underline{\em Obstacle 3:} Actively avoiding samples too close to original gradient.}
This is achieved by sampling perturbations $\{G^1, G^2, \ldots, G^T\}$ that are provably orthogonal to the true gradients at $\theta_\tau$: $\{\nabla \ell (f_{\theta_{\tau}}(X_{j,k}), Y_{j,k})\}_{j}$.
We obtain $\{G^1, G^2, \ldots, G^T\}$ by projecting $\{G'^1,G'^2,\ldots,G'^T\}$ onto the orthogonal subspace of the original gradients of client $k$. More precisely, we project them into the orthogonal subspace 
    \begin{equation}
    \begin{aligned}
  \!\!\!  S_k^\perp = \text{span} &\{ \forall w \in V :  \langle w, \nabla l(f(X_{j,k}, \theta)),c) \rangle  =  0, \\
    &\forall (X_{j,k}, \cdot)\in D^k, \forall c \in [1, C] \},
        \end{aligned}
\end{equation}
where $\langle \cdot,\cdot\rangle$ is the inner product, and $S_k^\perp$ is 
obtained through the Gram-Schmidt algorithm. An \textit{orthogonal subspace} can be understood as a vector space where each vector is perpendicular to the vectors in another subspace~\cite{orthogonal_conlr, continual_learn, orth_subspace}.
For example, in an $m$-dimensional space represented by \(x_1, \ldots , x_m\), if one subspace consists of vectors along the \(x_1\)-axis, an orthogonal subspace might be confined to the \(x_2,\ldots,x_m\)-plane. 
This arrangement ensures that changes or optimizations made within one subspace (along the \(x_1\)-axis) have no direct impact on the vectors lying in the orthogonal subspace (within the \(x_2,\ldots,x_m\)-plane).

\smallskip \noindent
\textbf{\underline{Complete solution:}} From the sampled set $\{G^1, G^2, \ldots, G^T\}$ of the orthogonal perturbations, we select the optimal $G^\star = \argmax_{G \in \{G^1,\ldots,G^T\}} P(D^k | \theta_\tau + G)$. By integrating this selection with orthogonal gradients, our method ensures both effective loss minimization and safeguards against gradient inversion attacks, protecting the client's private data from potential breaches.
The pseudocode of our complete algorithm \Tech{} is shown in \Cref{algo:our_tech}, with more details covered in the next few paragraphs.

\begin{algorithm}[ht]
\small
\caption{Pseudocode of \Tech{}}
\label{algo:our_tech}
\begin{algorithmic}[1]
    \State \textbf{Global Server Input:} Global model function $f$, the learning rate $\eta$, total FL training round $R$, and a set of randomly selected clients $\{1, 2, \cdots, K\}$ at round $\tau$
    \State \textbf{Local Client Input:} The $k$-th client's local dataset $D^k$, where $1 \leq k \leq K$, and the global model parameters $\theta_{\tau}$ at round $\tau$
    \For{\texttt{each training round $\tau$ in $\{1, 2, \cdots, R\}$}}
        \For{\texttt{each client $k$ in $\{1, 2, \cdots, K\}$}}
            \State $G_{\tau+1}^{k} \gets$ \texttt{\MakeUppercase{Local\_Update}}($\theta_{\tau}, D^{k}$)
            \Comment{$k$-th client trains on her local data and submits the update to the global server}
        \EndFor
        \State $G_{\tau+1} = \frac{1}{K} \sum_{k = 1}^{K}G_{\tau+1}^{k}$ 
        \Comment{Global server aggregates the received gradients from local clients}
        \State $\theta_{\tau+1} = \theta_{\tau} - \eta \cdot G_{\tau+1}$
        \Comment{Update of the global model parameters}
    \EndFor

    \Function{\texttt{\MakeUppercase{Local\_Update}}}{$\theta_{\tau}, D^{k}$}
        \State $G^0 = \nabla_{\theta}f(D^{k}, \theta_{\tau})$
        \Comment{Derive the original gradient}
        \State $\ell^{\star} \gets \texttt{\MakeUppercase{Evaluate}}(f_{\theta_{\tau}}, D^{k})$
        \Comment{Initialize the best loss}
        \State $G^{\star} = G^0$
        \Comment{Initialize the best gradient}
        
        \For {\texttt{each trial $t$ in $T$}} \Comment{$T$ is the number of trials}
            \State $G^t \gets \texttt{\MakeUppercase{Orthogonal\_Grad}}(G^0)$
            \Comment{\textbf{Phase 1}}
            \State $G^t_N \gets \texttt{\MakeUppercase{Normalize\_Grad}}(G^t, G^0)$
            \Comment{\textbf{Phase 2}}
            \State $\theta^t = \theta_{\tau} - \eta \cdot G^t_N$
            \Comment{\textbf{Phase 3}}
            \State $\ell^t = \texttt{\MakeUppercase{Evaluate}}(f_{\theta^t}, D^{k})$
            \Comment{Evaluate the effect of the current gradient}
            \If{$\ell^t < \ell^{\star}$}
            \Comment{Update the best loss and gradient}
                \State $\ell^{\star} = \ell^t$
                \State $G^{\star} = G^t_N$
            \EndIf
        \EndFor
        \State \Return $G^{\star}$
    \EndFunction
    
    \Function{\texttt{\MakeUppercase{Evaluate}}}{$f_{\theta}, D^k: \{x, y\}$}
        \State $\ell = \mathcal{L}(f_{\theta}(x), y)$
        \Comment{$\mathcal{L}$ is the loss function, e.g., Cross Entropy}
        \State \Return $\ell$
    \EndFunction
    
    \Function{\texttt{\MakeUppercase{Orthogonal\_Grad}}}{$G^0$}
        \State $G^t = \{\}$
        \For{$g_{l}$ in $G^0$}
        \Comment{Layer-wise orthogonal gradient projection}
            \State Sample $g_r \sim \mathcal{N}$
            \Comment{$\mathcal{N}$ denotes a normal distribution}
            \State $g_l^o = g_r - \text{proj}_{g_{l}}(g_r)$ \Comment{Project $g_r$ onto $g_{l}$}
            \State $G^t = G^t \cup \{g_l^o\}$ \Comment{Append layer-wise gradient back to the gradient set}
        \EndFor
        \State \Return $G^t$
    \EndFunction
    
    \Function{\texttt{\MakeUppercase{Normalize\_Grad}}}{$G^t, G^0$}
    \State $G^t_N = \{\}$
    \For{each pair $(g_l^o, g_l)$ in ($G^t, G^0$)}
    \Comment{Layer-wise operation}
        \State $\Tilde{g}_l^o = g_l^o \cdot \frac{\norm{g_l}_2}{\norm{g_l^o}_2}$
        \Comment{Normalize $g_l^o$}
        \State $G^t_N = G^t_N \cup \{\Tilde{g}_l^o\}$
    \EndFor
    \State \Return $G^t_N$
\EndFunction

\end{algorithmic}
\end{algorithm}

\begin{figure*}[ht!]
    \centering
    \begin{minipage}[t]{.74\linewidth}
        \centering
        \includegraphics[width=1\textwidth]{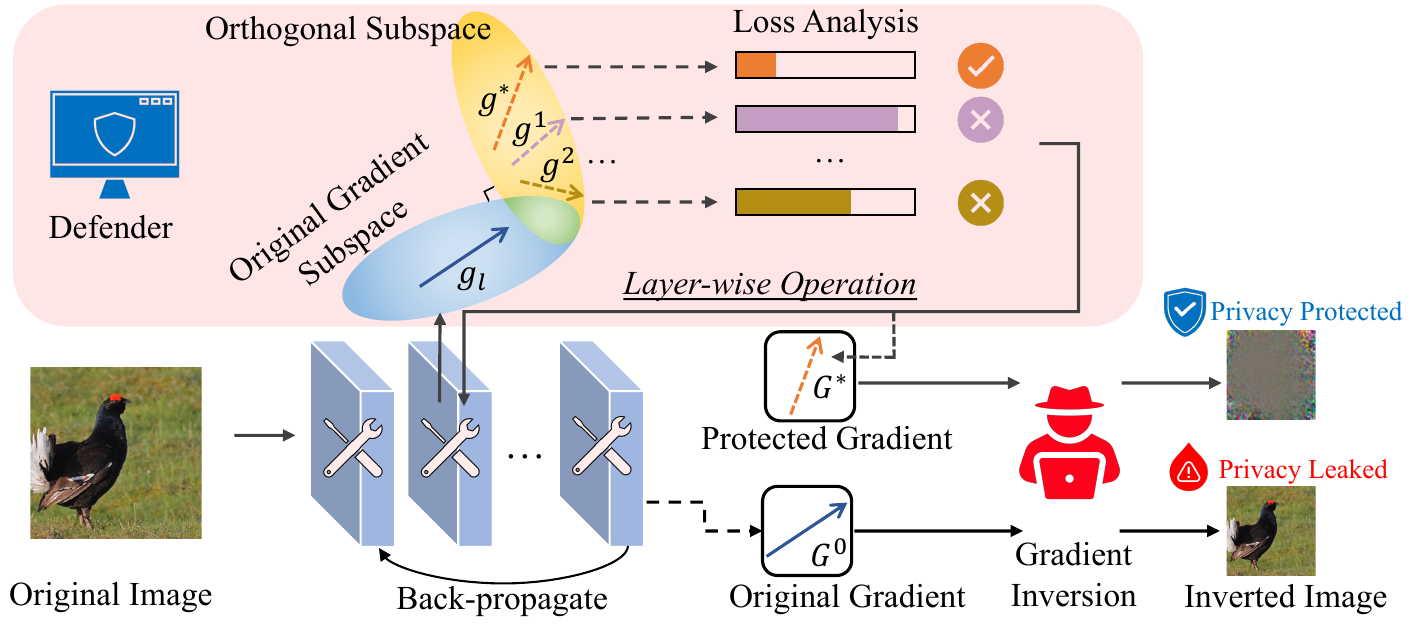}
        \caption{Overview of \Tech{}.}
        \label{fig:overview}
    \end{minipage}
    \hfill
    \nextfloat
    \begin{minipage}[t]{.25\linewidth}
        \centering
        \includegraphics[width=0.8\textwidth]{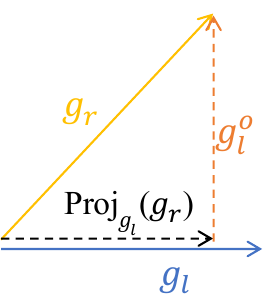}
        \caption{Orthogonal projection.}
        \label{fig:projection}
    \end{minipage}
\end{figure*}

\subsection{\Tech{} Details}

Consider a benign local client. She trains a model on her local dataset and achieves a gradient vector that minimizes the training loss.
She then submits her gradient to the honest-but-curious global server, the global server may invert her raw training data from the submitted gradient.
We address the scenario where the benign local client aims to protect the privacy of local training data without significantly reducing model accuracy.
In high-level, \Tech{} samples gradients in a \textit{subspace} that is \textit{orthogonal} to the original gradient layer by layer and select the one that achieves the lowest loss.
The overview of \Tech{} is illustrated in Figure~\ref{fig:overview}.
\Tech{} conducts defense to safeguard local data privacy, layer by layer. Beginning from the left and moving towards the right of Figure~\ref{fig:overview}, the local client computes the original gradient of the input image through back-propagation. \Tech{} then transforms the original gradients into protected ones, as depicted in the highlighted pink area. Specifically, \Tech{} generates a set of new gradients that are orthogonal to the original one. Subsequently, it undergoes a loss analysis to select the gradient that minimizes the training loss, and returns the optimal gradient. While an attacker may invert the original image based on the original gradient, it fails to produce a meaningful image for the protected gradient.

In addition to the overview, we offer a detailed description of our approach in Algorithm~\ref{algo:our_tech}, outlined below.

\smallskip \noindent
\textbf{FL Training Paradigm.}
We formally define the typical FL training paradigm in Line 1-7.
Line 1 specifies the input of the global server where $f$ denotes the global model function, $\eta$ is the learning rate, and $R$ is the total number of training rounds. At each round $\tau$, the global model will randomly select a set of $K$ clients for training.
Line 2 introduces the local client's input. It has its own local dataset $D^k$ and optimizes the global model parameter $\theta_{\tau}$ at round $\tau$.
In Line 3-5, each local client $k$ derives its local update based on the local data for each round $\tau$. Function \texttt{\MakeUppercase{Local\_Update}} specifies this procedure.
It then submits the local update to the global server and the global server aggregates the received gradients in Line 6.
Finally, the global server update the global model parameters in Line 7.

\smallskip \noindent
\textbf{\Tech{}'s Calculation of Local Updates.}
\Tech{} mitigates the local data privacy leakage and operates on the benign local clients.
Line 8-20 describe \Tech{}'s calculation of the local update based on the current global model parameters and the local data.
It consists of 3 principle phases: (1) Obtaining layer-wise orthogonal gradient updates, (2) Normalization, and (3) Selecting the best gradient according to the loss decrease.
Before delving into the main phases, \Tech{} first derives the original gradient without defense in Line 9, and initializes the best loss in Line 10.
Function \texttt{\MakeUppercase{Evaluate}}, detailed in Line 21-23, specifies how to calculate the initial loss.
In Line 11, \Tech{} initializes the best gradient using the original one.
Line 12-19 detail the main process. To obtain a best local update, \Tech{} takes a number of trials $T$ and selects the best gradient for submission. Usually $T=20$ is sufficient. We perform ablation study on the number of trials in Section~\ref{subsec:ablation_num_of_trials}.

\smallskip \noindent
\textbf{Phase 1: Layer-wise Orthogonal Gradient Update.}
To prevent gradient inversion, \Tech{} produces gradient update that is orthogonal to the original one.
Line 13 denotes the procedure, where Function \texttt{\MakeUppercase{Orthogonal\_grad}} takes the original gradient $G^0$ and returns a orthogonal one, layer by layer.
Details of the function are presented in Line 24-30.
\Tech{} initializes the gradient for the entire model in Line 25.
In Line 26, \Tech{} takes the gradient $g_l$ in each layer.
It then samples a new gradient $g_r$ with the same shape of $g_l$ from a normal distribution $\mathcal{N}$ in Line 27.
To ensure the orthogonality, \Tech{} projects $g_r$ on $g_l$ in Line 28 and append the orthogonal fraction to $G^t$ in Line 29.

The projection is illustrated in Figure~\ref{fig:projection}, where \Tech{} derives the orthogonal projection by subtracting the component of $g_r$ that aligns with $g_l$'s direction (Proj$_{g_l}(g_r)$), and then derives the orthogonal component $g_l^o$.
In addition, we introduce a formalization to clarify the projection process utilized in the preceding steps. Specifically, the projection of vector $g_r$ in the direction of vector $g_{l}$ is defined as $\text{proj}_{g_{l}}(g_r) = \frac{\langle g_r, g_{l} \rangle}{\langle g_{l}, g_{l} \rangle}g_{l}$. Applying this formalism, the orthogonal gradient for layer $l$, $g_l^o$, is computed as:

\begin{equation}
g_l^o = g_r - \text{proj}_{g_{l}}(g_r) = g_r - \frac{\langle g_r, g_{l} \rangle}{\langle g_{l}, g_{l} \rangle}g_{l}.
\end{equation}

\smallskip \noindent
\textbf{Phase 2: Normalization.}
Besides orthogonal projection, \Tech{} incorporates layer-wise normalization according to the original scale, aiming to avoid some of the gradient matrix including significant large values.
Line 14 presents the normalization phase, which is elaborated on in Lines 31-36.
\Tech{} initializes the returning gradient of the entire model in Line 32.
For the orthogonal gradient $g_l^o$ and the original one $g_l$ in each layer, \Tech{} normalizes $g_l^o$ to the scale of $g_l$ and derives $\Tilde{g}_l^o$ in Lines 33-34.
It then appends the normalized layer-wise gradient to $G^t_N$ in Line 35.

\smallskip \noindent
\textbf{Phase 3: Gradient Selection.}
To ensure a positive local update to the global model, \Tech{} selects the best normalized orthogonal gradient in all trials, according to their contribution to the loss reduction.
The selection process is presented in Line 15-19.
In Line 15, \Tech{} applies a gradient candidate $G^t_N$ to the model and evaluate its loss in Line 16.
It then compares the current loss $\ell^t$ with the best loss $\ell^{\star}$ (Line 17), and updates the best loss $\ell^{\star}$ (Line 18) and the best gradient $G^{\star}$ (Line 19) if $\ell^t$ is smaller than $\ell^{\star}$.

Finally, the optimal gradient $G^{\star}$ is returned (Line 20) as the protected local update. \Tech{} guarantees the privacy of local data by employing orthogonal projection and maintains global model performance through normalization and loss-guided gradient selection.

\section{Evaluation}\label{sec:evaluation}

In this section, we provide a comprehensive empirical evaluation of our proposed defense \Tech{}. 
We outline the experimental setup in Section~\ref{subsec:exp_setup}. 
Section~\ref{subsec:all_sota_attacks_defenses} presents the assessment of \Tech{}'s defense effectiveness against five attacks and compares it with four state-of-the-art defenses, across different batch sizes. 
We perform a convergence study in \textit{non-i.i.d.} federated learning setting in Section~\ref{subsec:convergence_study}.
In Section~\ref{subsec:adaptive}, we examine \Tech{}'s effectiveness against adaptive attacks.
Section~\ref{subsec:ablation} provides several ablation studies to investigate the impact of different design components and hyper-parameters of \Tech{}.

\subsection{Experimental Setup} \label{subsec:exp_setup}
To validate \Tech{}'s defense performance, we conduct experiments using five state-of-the-art attacks: IG~\cite{ig_geiping}, GI~\cite{gi_yin}, GGL~\cite{ggl}, GIAS~\cite{gias}, and GIDF~\cite{gifd}. We compare \Tech{} with four defense baselines: Noise~\cite{dp_fl}, Clipping~\cite{clipping}, Sparsi~\cite{compression}, and Soteria~\cite{soteria}.
We following existing attacks to adopt a randomly initialized ResNet-18~\cite{resnet} as the FL initial model, and employ the negative cosine similarity as distance metric $\mathcal{D}(\cdot)$ to measure the discrepancy between the inverted gradient and the original version.
To ensure fairness, we strictly follow the official implementation of both the attacks and the baseline defenses.
Our evaluation covers three widely-used dataset: ImageNet~\cite{imagenet}, FFHQ~\cite{ffhq} (10-class, using age as label) and CIFAR-10~\cite{cifar10}.
We use batch size $B = 1$ as the default setting, representing the easiest scenario for the attacker and most challenging for the defender.
The default number of sampling trials for \Tech{} is 20, and we conduct an ablation study on this parameter in Section~\ref{subsec:ablation_num_of_trials}.
Perturbed gradients are drawn from a normal distribution $\mathcal{N}$. Additionally, these gradients can also be constructed in alternative ways, such as from other meaningful images to mislead the attacker. We present the results of this experiment in Section~\ref{subsec:ablation}.
Code is available at \url{https://censor-gradient.github.io}.

\smallskip 
\ndssrevise{
\noindent
\textbf{Attacks Configurations.}
(1) Inverting Gradients (IG)~\cite{ig_geiping} uses Adam on signed gradients, with cosine similarity to optimize the input initialized from Gaussion;
(2) GradInversion (GI)~\cite{gi_yin} initializes the pixels from Gaussian noise and uses Adam to optimize them with gradient matching;
(3) Gradient Inversion in Alternative Spaces (GIAS)~\cite{gias} uses negative cosine as the gradient dissimilarity function and the same Adam is used as the optimizer;
and (4) Generative Gradient Leakage (GGL)~\cite{ggl} applies BigGAN~\cite{biggan} on ImageNet
and CIFAR-10, and StyleGAN2~\cite{ffhq} on FFHQ. They also use KL-based regularization and CMA-ES optimizer;
(5) Gradient Inversion over Feature Domains (GIFD)~\cite{gifd} uses intermediate features of BigGAN or StyleGAN2 and utilizes Adam optimizer with a warm-up strategy. They apply regularization with $\ell_2$ distance.

\smallskip 
\noindent
\textbf{Defenses Configurations.}
We follow the same defense setup as the previous work~\cite{gifd, ggl, gias}: 
(1) Gaussian Noise~\cite{dp_fl} randomly perturbs the gradients using Gaussian noises with a standard deviation of 0.1;
(2) Gradient Clipping~\cite{clipping} constrains the magnitude of gradients by clipping each value within a clipping bound;
(3) Gradient Sparsification~\cite{compression} maps small absolute gradients to zero and only transmits the largest values during update;
and (4) Soteria~\cite{soteria} prunes by applying a mask to the defended fully connected layer's gradients.
}

\smallskip
\noindent
\textbf{Evaluation Metrics.}
In addition to qualitative visual comparison, we utilize the following metrics for quantitatively evaluating the similarity between the inverted images and their original versions:
\begin{enumerate}
    \item \textbf{Mean Square Error (MSE ↑)}~\cite{dodge2008concise}. It calculates the pixel-wise MSE between the inverted images and the original images. Higher MSE indicates less similarity and stronger privacy protection;
    \item \textbf{Learned Perceptual Image Patch Similarity (LPIPS ↑)}~\cite{zhang2018unreasonable}. LPIPS measures the perceptual image similarity between the reconstructed image features and those of ground-truth images, measured by a pre-trained VGG network. Higher LPIPS values suggest better defense;
    \item \textbf{Peak Signal-to-Noise Ratio (PSNR ↓)}~\cite{lin2005visual}. PSNR is the ratio of the maximum squared pixel fluctuation between two images. Lower PSNR values indicate better protection;
    \item \textbf{Similarity Structural Index Measure (SSIM ↓)}~\cite{wang2003multiscale}. SSIM evaluates the perceptual similarity between two images, considering their luminance, contrast, and structure. Lower SSIM values indicate stronger privacy protection.
\end{enumerate}

\begin{table*}[t]
    \centering
    \fontsize{7.3}{12}\selectfont
    \tabcolsep=0.7pt
\caption{\ndssrevise{Quantitative evaluation of various defense methods against existing attacks. (An upward arrow denoting the higher the better, a downward arrow denoting the lower the better.)}}
\label{tab:bigtable}

\begin{tabular}{llccccggggccccggggcccc}
\dtoprule

\multirow{2}{*}{DA}  & \multicolumn{1}{c}{\multirow{2}{*}{Defense}} & \multicolumn{4}{c}{IG~\cite{ig_geiping}}                                                                                                 & \multicolumn{4}{c}{GI~\cite{gi_yin}}                                                                                                 & \multicolumn{4}{c}{GGL~\cite{ggl}}                                                                                                & \multicolumn{4}{c}{GIAS~\cite{gias}}                                                                                               & \multicolumn{4}{c}{GIFD~\cite{gifd}}                                                                                               \\
\cmidrule(lr){3-6} \cmidrule(lr){7-10}\cmidrule(lr){11-14}\cmidrule(lr){15-18}\cmidrule(lr){19-22}
                          & \multicolumn{1}{c}{}                         & \multicolumn{1}{c}{\textcolor{red}{MSE↑}} & \multicolumn{1}{c}{\textcolor{red}{LPIPS↑}} & \multicolumn{1}{c}{\textcolor{blue}{PSNR↓}} & \multicolumn{1}{c}{\textcolor{blue}{SSIM↓}} & \multicolumn{1}{g}{\textcolor{red}{MSE↑}} & \multicolumn{1}{g}{\textcolor{red}{LPIPS↑}} & \multicolumn{1}{g}{\textcolor{blue}{PSNR↓}} & \multicolumn{1}{g}{\textcolor{blue}{SSIM↓}} & \multicolumn{1}{c}{\textcolor{red}{MSE↑}} & \multicolumn{1}{c}{\textcolor{red}{LPIPS↑}} & \multicolumn{1}{c}{\textcolor{blue}{PSNR↓}} & \multicolumn{1}{c}{\textcolor{blue}{SSIM↓}} & \multicolumn{1}{g}{\textcolor{red}{MSE↑}} & \multicolumn{1}{g}{\textcolor{red}{LPIPS↑}} & \multicolumn{1}{g}{\textcolor{blue}{PSNR↓}} & \multicolumn{1}{g}{\textcolor{blue}{SSIM↓}} & \multicolumn{1}{c}{\textcolor{red}{MSE↑}} & \multicolumn{1}{c}{\textcolor{red}{LPIPS↑}} & \multicolumn{1}{c}{\textcolor{blue}{PSNR↓}} & \multicolumn{1}{c}{\textcolor{blue}{SSIM↓}} \\

\midrule

\multirow{6}{*}{{\rotatebox[origin=c]{90}{ImageNet}}} & No Defense                                   & 0.0195                    & 0.5574                            & 17.819                      & 0.2309                    & 0.0191                    & 0.5402                            & 17.908                      & 0.2400                    & 0.0453                    & 0.5952                            & 13.873                      & 0.0745                    & 0.0191                    & 0.4795                            & 18.452                      & 0.3099                    & 0.0130                    & 0.3782                            & 21.364                      & 0.4528                    \\
\cmidrule{2-22}
                          & Noise~\cite{dp_fl}                                           & 0.0246                    & 0.6294                            & 16.338                      & 0.1754                    & 0.0269                    & 0.6300                            & 15.883                      & 0.1549                    & 0.0410                    & 0.5697                            & 14.252                      & 0.0817                    & 0.0253                    & 0.5947                            & 16.601                      & 0.1854                    & 0.0196                    & 0.5380                            & 18.166                      & 0.2686                    \\
                          & Clipping~\cite{clipping}                                     & 0.0167                    & 0.5008                            &18.883                      & 0.3128                    & 0.0383                    & 0.7302                            & 14.844                      & 0.0106                    & \textbf{0.0477}           & 0.5823                            & \textbf{13.520}             & 0.0749                    & 0.0203                    & 0.4825                            & 18.738                      & 0.3186                    & 0.0150                    & 0.4433                            & 19.547                      & 0.3798                    \\
                          & Sparsi~\cite{compression}                  & 0.0137                    & 0.4945                            & 19.383                      & 0.3419                    & 0.0157                    & 0.4941                            & 18.799                      & 0.3099                    & 0.0456                    & 0.6080                            & 13.743                      & 0.0776                    & 0.0135                    & 0.3981                            & 20.483                      & 0.4182                    & 0.0179                    & 0.4444                            & 19.486                      & 0.3686                    \\
                          & Soteria~\cite{soteria}                                      & \textbf{0.0662}           & \textbf{0.7596}                   & \textbf{12.220}             & 0.0135                    & \textbf{0.0682}           & 0.7485                            & \textbf{12.215}             & 0.0134                    & 0.0461                    & 0.5986                            & 13.879                      & 0.0708                    & 0.0245                    & 0.4986                            & 17.646                      & 0.2664                    & 0.0139                    & 0.3967                            & 20.602                      & 0.4335                    \\
\cmidrule{2-22}                          
                          & \Tech{}                                         & 0.0600                    & 0.7551                            & 12.463                      & \textbf{0.0067}           & 0.0416                    & \textbf{0.8615}                   & 14.446                      & \textbf{0.0021}           & 0.0419                    & \textbf{0.7912}                   & 14.262                      & \textbf{0.0094}           & \textbf{0.0650}           & \textbf{0.7591}                   & \textbf{12.266}             & \textbf{0.0139}           & \textbf{0.0507}           & \textbf{0.7610}                   & \textbf{13.323}             & \textbf{0.0094}        \\

\midrule[1pt]

\multirow{6}{*}{{\rotatebox[origin=c]{90}{FFHQ}}}    & No Defense                                   & 0.0143                    & 0.5247                            & 18.666                      & 0.4209                    & 0.0194                    & 0.5692                            & 17.246                      & 0.3490                    & 0.0421                    & 0.5424                            & 14.167                      & 0.1953                    & 0.0173                    & 0.4228                            & 18.738                      & 0.4792                    & 0.0149                    & 0.4353                            & 20.116                      & 0.5102                    \\
\cmidrule{2-22}                          
                         & Noise~\cite{dp_fl}                                           & 0.0311                    & 0.6666                            & 15.298                      & 0.2377                    & 0.0388                    & 0.7131                            & 14.243                      & 0.1507                    & 0.0454                    & 0.5784                            & 13.866                      & 0.1752                    & 0.0225                    & 0.5878                            & 17.070                      & 0.3449                    & 0.0190                    & 0.5250                            & 17.953                      & 0.3972                    \\
                         & Clipping~\cite{clipping}                                     & 0.0262                    & 0.6245                            & 15.963                      & 0.2865                    & 0.0593                    & 0.8223                            & 12.514                      & 0.0080                    & 0.0381           & 0.4881                            & 14.571             & 0.2286                    & 0.0171                    & 0.4267                            & 18.693                      & 0.4808                    & 0.0147                    & 0.4353                            & 20.113                      & 0.5139                    \\
                         & Sparsi~\cite{compression}                  & 0.0225                    & 0.5745                            & 17.113                      & 0.3411                    & 0.0278                    & 0.6189                            & 16.068                      & 0.2889                    & 0.0421                    & 0.5479                            & 14.182                      & 0.2035                    & 0.0146                    & 0.4057                            & 19.490                      & 0.5130                    & 0.0125                    & 0.4070                            & 20.581                      & 0.5423                    \\
                         & Soteria~\cite{soteria}                                      & \textbf{0.1124}           & \textbf{0.8446}                   & \textbf{9.662}              & 0.0165                    & \textbf{0.1105}           & \textbf{0.8467}                   & \textbf{9.753}              & 0.0157                    & 0.0438                    & 0.5279                            & 13.842                      & 0.2014                    & 0.0114                    & 0.3560                            & 20.551                      & 0.5489                    & 0.0165                    & 0.4686                            & 19.296                      & 0.4789                    \\
\cmidrule{2-22} 
                         & \Tech{}                                        & 0.1009                    & 0.8347                            & 10.125                      & \textbf{0.0080}           & 0.0992                    & 0.8335                   & 10.550                      & \textbf{0.0061}           & \textbf{0.0823}           & \textbf{0.8155}                   & \textbf{10.947}             & \textbf{0.0219}           & \textbf{0.1108}           & \textbf{0.7895}                   & \textbf{9.613}              & \textbf{0.0268}           & \textbf{0.1037}           & \textbf{0.8097}                   & \textbf{9.904}              & \textbf{0.0195}          

\\

\midrule[1pt]

\multirow{6}{*}{{\rotatebox[origin=c]{90}{CIFAR-10}}}    & No Defense                                   & 0.0023                    & 0.0905                            & 26.324                      & 0.8139                    & 0.0192                 & 0.3909                            & 17.170                      & 0.4574                    & 0.0275                    & 0.5569                            & 15.601                      & 0.1099                    & 0.0009                    & 0.0333                            & 30.414                      & 0.9276                    & 0.0201                    & 0.5297                            & 16.968                      & 0.2408                    \\
\cmidrule{2-22}                          
                         & Noise~\cite{dp_fl}                                           & 0.0052                    & 0.2229                            & 22.810                      & 0.7010                    & 0.0060                    & 0.2587                            & 22.242                      & 0.7241                    & 0.0292                    & 0.5974                            & 15.339                      & 0.1002                    & 0.0022                    & 0.2600                            & 26.660                      & 0.6555                    & 0.0137                    & 0.5166                            & 18.648                      & 0.3581                    \\
                         & Clipping~\cite{clipping}                                     & 0.0036                    & 0.1050                            & 24.396                      & 0.8048                    & 0.0496                    & 0.5317                            & 13.042                      & 0.0482                    & \textbf{0.0689}           & 0.6125                            & \textbf{11.621}             & 0.0935                    & 0.0014                    & 0.0568                            & 28.567                      & 0.8863                    & 0.0228                    & 0.5492                            & 16.426                      & 0.2179                    \\
                         & Sparsi~\cite{compression}                  & 0.0093                    & 0.3210                            & 20.311                      & 0.5933                    & 0.0041                    & 0.1631                            & 23.841                      & 0.7614                    & 0.0583                    & 0.6491                            & 12.340                      & 0.1407                    & 0.0003                    & 0.0100                            & 35.652                      & 0.9631                    & 0.0297                    & 0.4557                            & 15.267                      & 0.2432                    \\
                         & Soteria~\cite{soteria}                                      & 0.0558           & 0.5846                   & 12.537              & 0.1222                    & \textbf{0.0891}           & 0.6986                   & \textbf{10.501}              & 0.0378                    & 0.0326                    & 0.6274                            & 14.867                      & 0.0731                    & 0.0012                    & 0.0461                            & 29.308                      & 0.9050                    & 0.0239                    & 0.5577                            & 16.222                      & 0.3907                    \\
\cmidrule{2-22} 
                         & \Tech{}                                        & \textbf{0.0939}                    & \textbf{0.6884}                            & \textbf{10.272}                      & \textbf{0.0112}           & 0.0806                    & \textbf{0.7006}                   & 10.937                      & \textbf{0.0167}           & 0.0260           & \textbf{0.7159}                   & 15.857             & \textbf{0.0179}           & \textbf{0.0789}           & \textbf{0.6819}                   & \textbf{11.027}              & \textbf{0.0789}           & \textbf{0.0916}           & \textbf{0.6398}                   & \textbf{10.381}              & \textbf{0.1452}          

\\

\dbottomrule

\end{tabular}
\end{table*}

\begin{figure*}[ht!]
    \centering
    \begin{minipage}[c]{0.48\textwidth}
        \centering
        \includegraphics[width=.98\textwidth]{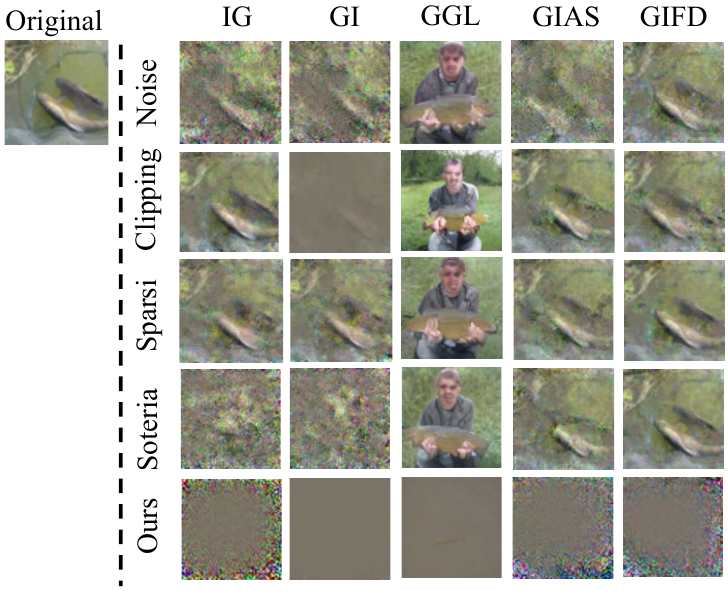}
    \end{minipage}
    \begin{minipage}[c]{0.48\textwidth}
        \centering
        \includegraphics[width=.98\textwidth]{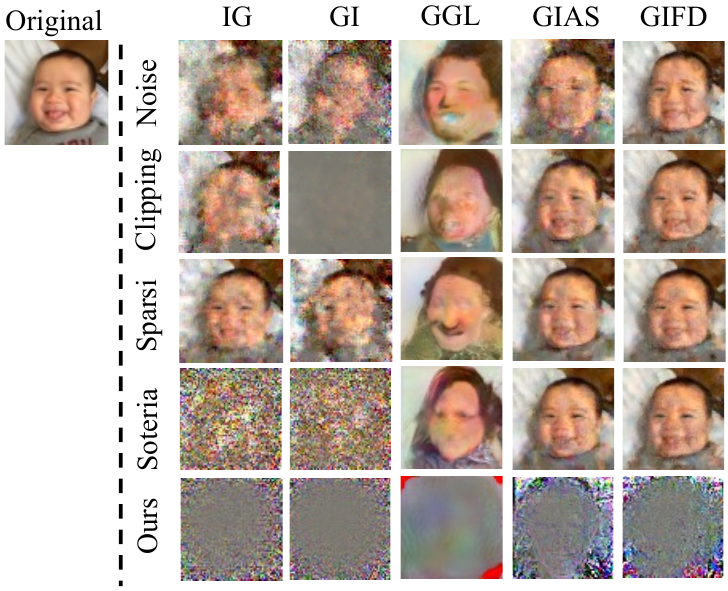}
    \end{minipage}
    \caption{Qualitative evaluation of various attack inversions under existing defenses.}
    \label{fig:qualitative}
\end{figure*}

\subsection{Comparison with State-of-the-art Defenses against Diverse Attacks}\label{subsec:all_sota_attacks_defenses}
In this section, we evaluate the effectiveness of our proposed defense, \Tech{}, against various gradient inversion attacks and compare its performance with several leading defense baselines. The quantitative results are presented in Table~\ref{tab:bigtable}. The first column of the table lists the datasets, the second column specifies the defense methods, and the subsequent columns present the performance of these defenses against five distinct attacks. Our evaluation includes three widely-used datasets and a typical model architecture, ResNet-18~\cite{resnet}, in line with prior works~\cite{ggl, gifd}. For each attack, we report four metrics, detailed in Section~\ref{subsec:exp_setup}. Here we report the initial round zero performance, as the start round is the easiest for the attacker to perform inversion and it is also the default setting in existing attack literatures~\cite{ig_geiping, gi_yin, ggl,gias, gifd}. These metrics provide a comprehensive assessment of inversion fidelity relative to the ground-truth images from the quantitative perspective.
We also provide qualitative results in Figure~\ref{fig:qualitative}, which visually illustrate the fidelity of the inverted images.
For certain attacks that require batch normalization (BN) statistics, it is important to note that in real-world FL systems, the BN statistics derived from private data are typically not transmitted. Consequently, we do not apply the strong BN prior for these attacks, following existing works official implementations~\cite{ggl, gifd}.
Given that the randomly initialized values of inverted images significantly impact the reconstruction outcomes, following existing work~\cite{gifd}, we perform four trials for each attack and report the attack inversion result with the best performance.
In addition, for each attack, we perform the gradient inversion on ten different images for every 1000-th image of ImageNet, FFHQ and CIFAR-10 validation set. We report the numerical results for quantitative metrics are the averages of 10 inversions. 
\ndssrevise{We also perform an overhead evaluation in Appendix~\ref{appen:overhead}.}

\smallskip \noindent
\textbf{Performance Against Stochastic Optimization Attacks.}
IG~\cite{ig_geiping} and GI~\cite{gi_yin} are two typical attacks that utilize stochastic optimization without the help of GANs. It is noteworthy that our defense, \Tech{}, achieves performance comparable to the state-of-the-art defense Soteria~\cite{soteria}, and significantly surpasses other defenses. The slight superiority (ranging from 2\%-18\% in various metrics) of Soteria over \Tech{} can be attributed to its special design of perturbed gradient.
Essentially, Soteria employs stochastic optimization to invert a dummy input, utilizing the gradient computed on the dummy input to deceive the inversion process. If it can accurately approximate the attacker's inverted image and prevent it in the gradient space, it achieves success in data protection. Since Soteria is based on stochastic optimization, it is highly effective against attacks relying on stochastic optimization, such as IG and GI. However, it may fall short in GAN-based inversion scenarios, as it cannot approximate the inversion process of GANs.
\Tech{}, though without such a dummy input inversion, still achieves comparable quantitative results and effectively prevents attackers from inverting meaningful images, as demonstrated in the last row of Figure~\ref{fig:qualitative}.
Moreover, Soteria is less effective against more sophisticated GAN-based attacks, e.g., GGL, GIAS and GIFD.
\Tech{}, on the other hand, maintains its efficacy across a broader range of attacks.

\smallskip \noindent
\textbf{Performance Against GAN-based Reconstruction Attacks.}
GAN-based attacks, i.e., GGL~\cite{ggl}, GIAS~\cite{gias} and GIFD~\cite{gifd}, typically produce better reconstruction results compared to stochastic optimization.
This is reasonable, as GAN-based inversion is facilitated by the pre-trained GAN to generate higher quality images.
From the quantitative results in Table~\ref{tab:bigtable}, we observe that \Tech{} outperforms existing defenses in almost all cases, and significantly surpasses the state-of-the-art defense Soteria (up to 114\% in the metrics).
We also observe that in few special cases, i.e., GGL attack on ImageNet and CIFAR-10 datasets, Clipping~\cite{clipping} slightly outperforms \Tech{} for 14\% in MSE loss and 5\% in PSNR.
This could potentially be reasoned as sometimes, large values in the gradient strongly reflect the data information from the original training samples, particularly in the case of GGL. In such scenarios, Clipping, which removes large values, can effectively and precisely safeguard privacy.
However, such cases are rare and \Tech{} is generally more effective than Clipping in all other cases.
Moreover, from the visualization in the last row of Figure~\ref{fig:qualitative}, we can observe even with the aid of GAN, it is challenging for the attacker to invert meaningful images under \Tech{}.

\begin{table}[t]
    \centering
    \fontsize{7}{12}\selectfont
    \tabcolsep=0.6pt
    \caption{Evaluation for batch size of 4.}
    \label{tab:larger_batch}
\begin{tabular}{llrrrrGGGG}
\toprule
\multirow{2}{*}{\textbf{DA}} & \multicolumn{1}{c}{\multirow{2}{*}{\textbf{Defense}}} & \multicolumn{4}{c}{\textbf{GIAS}}                                                                                                                           & \multicolumn{4}{c}{\textbf{GIFD}}                                                                                                                           \\
\cmidrule(lr){3-6} \cmidrule(lr){7-10}
                                  & \multicolumn{1}{c}{}                                  & \multicolumn{1}{c}{\textbf{MSE ↑}} & \multicolumn{1}{c}{\textbf{LPIPS ↑}} & \multicolumn{1}{c}{\textbf{PSNR ↓}} & \multicolumn{1}{c}{\textbf{SSIM ↓}} & \multicolumn{1}{g}{\textbf{MSE ↑}} & \multicolumn{1}{g}{\textbf{LPIPS ↑}} & \multicolumn{1}{g}{\textbf{PSNR ↓}} & \multicolumn{1}{g}{\textbf{SSIM ↓}} \\
\midrule
\multirow{6}{*}{{\rotatebox[origin=c]{90}{ImageNet}}}           & No Defense                                            & 0.0249                             & 0.5569                                     & 16.519                              & 0.2096                              & 0.0271                             & 0.5582                                     & 16.332                              & 0.2309                              \\
\cmidrule{2-10}
                                  & Noise                                                 & 0.0368                             & 0.6551                                     & 14.776                              & 0.1077                              & 0.0350                             & 0.6401                                     & 14.991                              & 0.1338                              \\
                                  & Clipping                                              & 0.0267                             & 0.5666                                     & 16.223                              & 0.2100                              & 0.0294                             & 0.5655                                     & 15.943                              & 0.2092                              \\
                                  & Sparsi                           & 0.0284                             & 0.5650                                     & 16.137                              & 0.2128                              & 0.0286                             & 0.5587                                     & 16.256                              & 0.2221                              \\
                                  & Soteria                                               & 0.0259                             & 0.5641                                     & 16.536                              & 0.2037                              & 0.0284                             & 0.5593                                     & 16.174                              & 0.2247                              \\
\cmidrule{2-10}
                                  & \Tech{}                                                  & \textbf{0.0776}                    & \textbf{0.7680}                            & \textbf{11.686}                     & \textbf{0.0099}                     & \textbf{0.0686}                    & \textbf{0.7614}                            & \textbf{12.245}                     & \textbf{0.0088}                                     
                                  \\
\bottomrule
\end{tabular}
\end{table}

\smallskip \noindent
\textbf{Performance of Larger Batch Sizes.}
Although a batch size of 1 at each local step is the simplest setting for attackers, we evaluate the defense performance with larger (4) batch sizes. Our experiments are conducted on ImageNet using two state-of-the-art attacks, i.e., GIAS and GIFD. Notably, for these attacks, we ensure that no duplicate labels are present in each batch and they infer the labels from the received gradients~\cite{gi_yin}.
We compare the defense performance of \Tech{} against other baselines. The results are presented in Table~\ref{tab:larger_batch}.
Observe that \Tech{} is robust against larger batch sizes and outperforms all other baselines across all metrics, indicating its general effectiveness.
We also study different number of training rounds in Appendix~\ref{appen:addi_1} and different number of clients in Appendix~\ref{appen:num_of_clients}.

\subsection{Convergence Study}\label{subsec:convergence_study}

We undertake a convergence study within the framework of federated learning as delineated by \cite{fedavg}.
\ndssrevise{
Our experimental evaluations are carried out using the CIFAR-10 dataset \cite{cifar10} and employ the ResNet-18 model architecture \cite{resnet}. Notably, our data are not independent and identically distributed (\textit{non-i.i.d.}), better reflecting the complexities of real-world scenarios. 
}
This experimental design is consistent with the approach of \cite{bagdasaryan2020backdoor}, which utilizes a Dirichlet distribution \cite{minka2000estimating} to simulate the \textit{non-i.i.d.} nature of the data.

\begin{figure}[t]
    \centering
    \includegraphics[width=.48\textwidth]{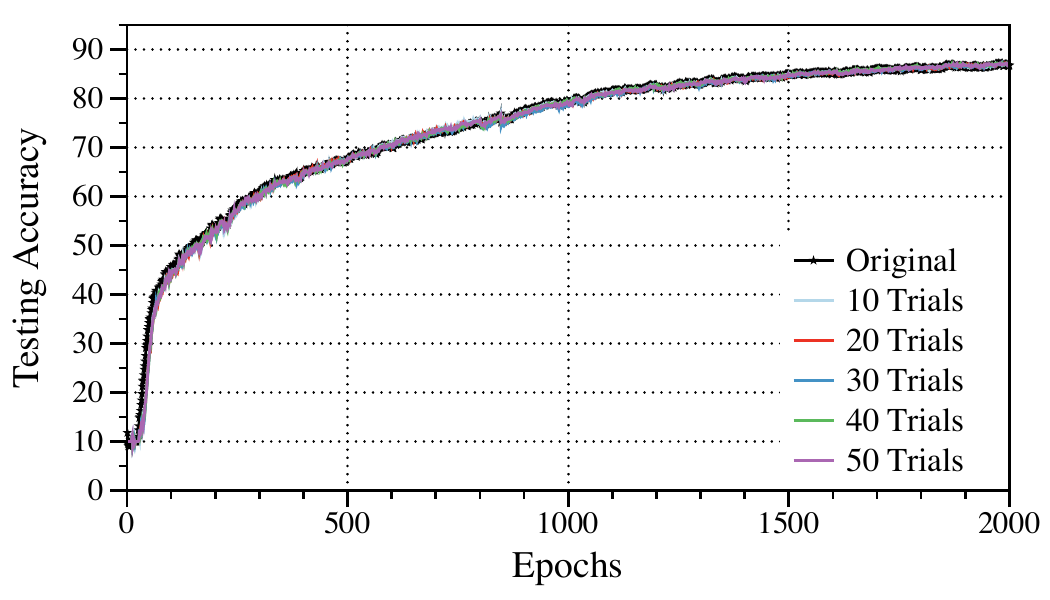}
    \caption{\ndssrevise{Testing accuracy on CIFAR-10.}}
    \label{fig:converge_acc}
\end{figure}

\begin{figure}[t]
    \centering
    \includegraphics[width=.48\textwidth]{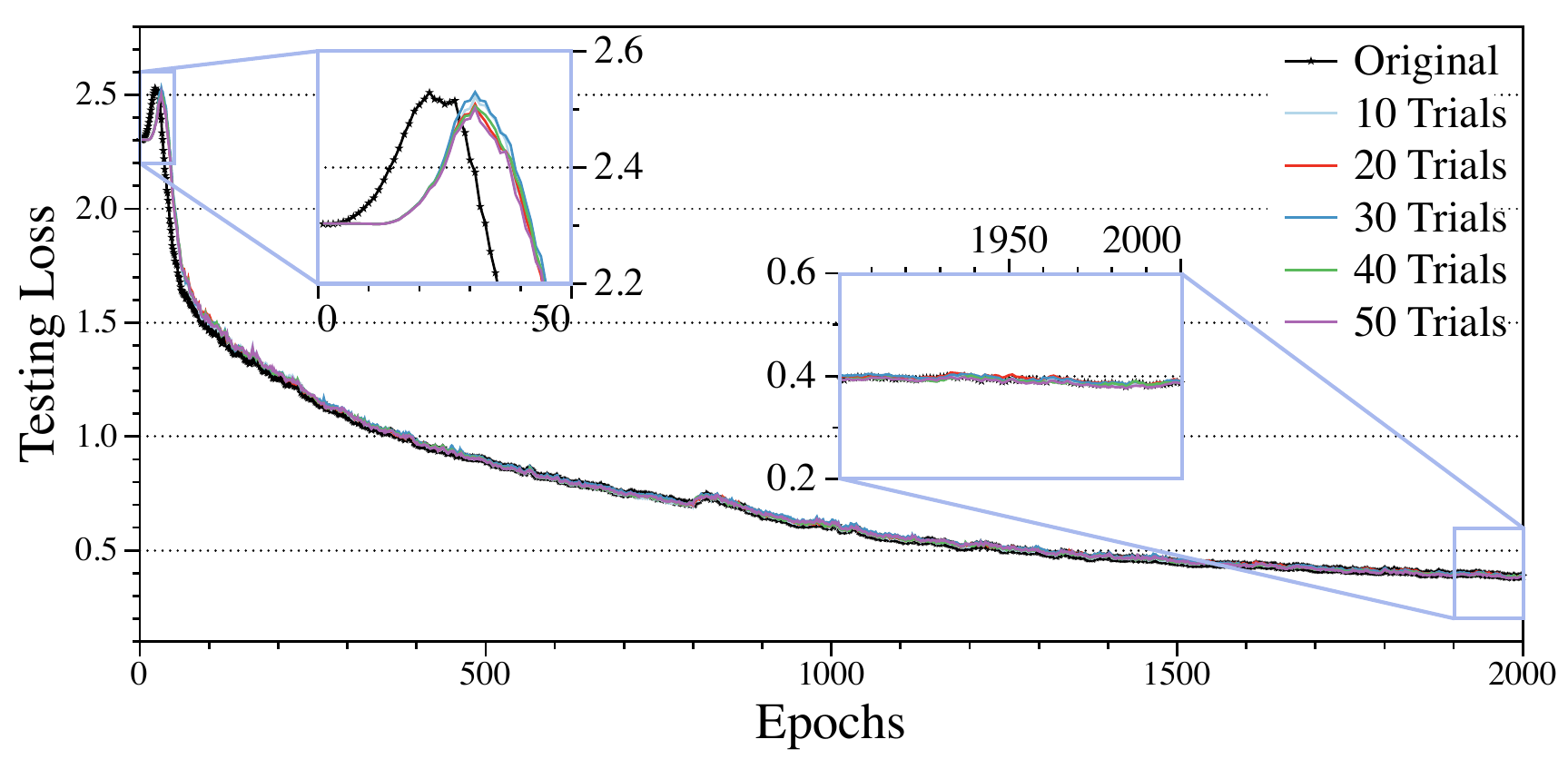}
    \caption{\ndssrevise{Testing loss on CIFAR-10.}}
    \label{fig:converge_loss}
\end{figure}

\ndssrevise{
Our experimental setup involves 100 clients by default, participating in a comprehensive training process spanning 2000 rounds. In each round, we randomly selected 10 clients to participate. 
}
The training utilizes stochastic gradient descent as the optimization technique, with a local learning rate set at 0.1 and a batch size of 64. Additionally, orthogonal gradient updates are implemented on each local client, and we assess the impact of different numbers of trials on the orthogonal gradient selection process.
\ndssrevise{
Figure~\ref{fig:converge_acc} displays the testing accuracy results. 
The testing accuracy across different experimental conditions reach roughly similar levels, with both standard original (vanilla) setting and adapted \Tech{}'s defense configurations achieving about 86\% accuracy after 2000 training rounds, on the CIFAR-10 dataset under \textit{non-i.i.d.} conditions. 
}
Notably, the discrepancies are minimal during the initial phases of training when comparing the vanilla setting with \Tech{}'s varying trial numbers setting. 
\ndssrevise{
Similarly, as illustrated in Figure~\ref{fig:converge_loss}, the testing loss exhibits only slight variations between the standard original training and \Tech{} defense strategy during the early stages of training.
By the end of the 2000-th round, the testing loss for both configurations have settled at a relatively low level.
The results indicate that \Tech{} will not impact the convergence of FL model training.
}

\subsection{Adaptive Attack: Expectation Over Transformation (EOT)}\label{subsec:adaptive}

As attackers may devise adaptive strategies to overcome \Tech{}, this section introduces a countermeasure and thoroughly evaluates \Tech{} under scenarios involving adaptive attacks. 
Our defense method, \Tech{}, is particularly designed to explore lower-dimensional manifolds within a high-dimensional neural network’s parameter space, it would be very challenging for the adversary to identify and map the gradient that is applied with defense to the original gradient.
To remedy this situation and construct strong adversary, we integrate state-of-the-art existing attack GIFD with the Expectation Over Transformation (EOT)~\cite{eot},
since EOT has been considered as an effective strategy to mitigate the random effect induced by the defenders~\cite{athalye2018obfuscated,tramer2020adaptive}.
The idea of EOT is to perform the gradient transformation multiple times, and take the average gradient over several runs, to approximate the gradient and mitigate the randomization effect as much as possible.

Our assessment is carried out using two well-known datasets, ImageNet and FFHQ, applying EOT alongside the state-of-the-art GIFD attack~\cite{gifd}. 
The attacker follows the sampling methodology used by \Tech{}'s defense algorithm, generating orthogonal gradients and subsequently normalizing them in the same manner.
After that, the attacker computes the average of the collected gradients over the number of trials.
During this procedure, the attacker also refine the averaging process to prioritize gradients that are more informative or indicative of the original data.
As demonstrated in Table~\ref{tab:eot}, even when subjected to the EOT-enhanced attack, \Tech{}'s performance remains unaffected, consistently producing significantly different inverted images compared to the original images.
This indicates that our defense mechanism still prevents the attacker to invert any useful information comparing without applying EOT.
This effectiveness distinctly highlights the robustness of \Tech{} against adaptive attacks.
The underlying reason is as discussed in the theoretical analysis in Section~\ref{sec:method} that since the orthogonal subspace of a gradient of a model with $m$ parameters has $(m - 1)$-dimensions, and the resulting orthogonal gradient is chosen from that high dimension, it would be very challenging for the adversary to identify and map the after-defense gradient to the original gradient.
\ndssrevise{
We also discuss other possible adaptive attacks in Appendix~\ref{appen:adaptive}.
}

\begin{table}[t]
    \centering
    \tabcolsep=7pt
    \caption{Adaptive attack with EOT.}
    \label{tab:eot}
    \begin{tabular}{llcccc}
    \toprule
    \textbf{Dataset}          & \textbf{EOT} & \multicolumn{1}{l}{\textbf{MSE ↑}} & \multicolumn{1}{l}{\textbf{LPIPS ↑}} & \multicolumn{1}{l}{\textbf{PSNR ↓}} & \multicolumn{1}{l}{\textbf{SSIM ↓}} \\
    \midrule
    \multirow{2}{*}{ImageNet}  & w/o & 0.0507         & 0.7610                 & 13.32           & 0.0094         \\
                               & w/.   & 0.0518         & 0.7668                 & 13.39           & 0.0087         \\
    \midrule
    \multirow{2}{*}{FFHQ}      & w/o & 0.1037         & 0.8097                 & 9.90            & 0.0195         \\
                               & w/.   & 0.1098         & 0.8340                 & 9.82            & 0.0195   
                                \\
    \bottomrule
    \end{tabular}
\end{table}

\subsection{Ablation Study}\label{subsec:ablation}
In this section, we conduct several ablation studies to investigate the impact of our design components and hyper-parameters. The effect of applying layer-wise operation can be found in Appendix~\ref{appen:abalation}.

\begin{figure}[t]
    \centering
    \includegraphics[width=.48\textwidth]{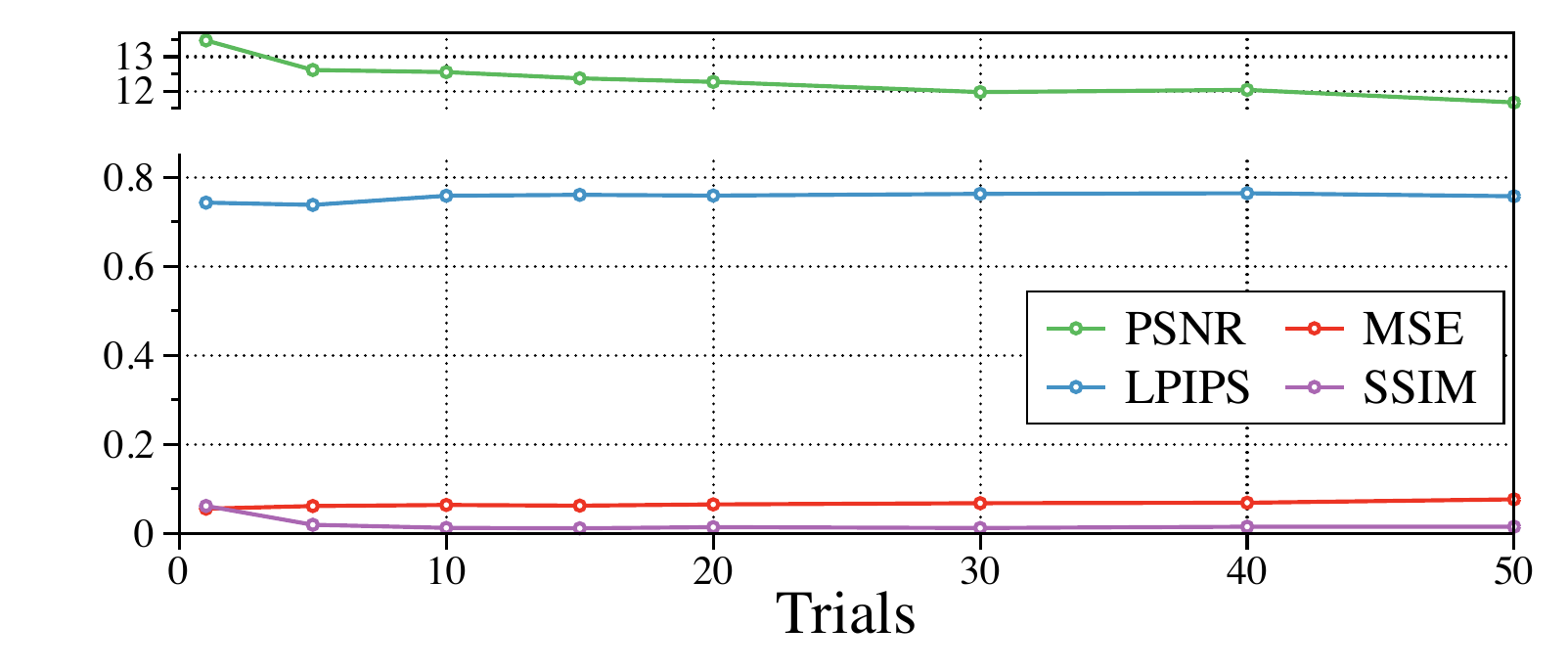}
    \caption{Different trials on ImageNet with GIAS.}
    \label{fig:ablation_trials}
\end{figure}

\smallskip \noindent
\textbf{Effect of Different Number of Trials.}\label{subsec:ablation_num_of_trials}
In Section~\ref{sec:method}, we mention that \Tech{} randomly samples $T = 20$ gradients and selects the one with the highest loss reduction. In this study, we investigate the impact of varying the number of trials. Our experiment is conducted on ImageNet using the GIAS attack under our defense.
Figure~\ref{fig:ablation_trials} shows the quantitative results across four metrics. Figure~\ref{fig:ablation_trials_example} (in Appendix~\ref{appen:abalation}) illustrates the inverted images for different numbers of trials.
We observe that initially the metrics tend to improve with an increasing number of trials, which is expected as more trials allow \Tech{} to find a more optimal gradient update. However, as the number of trials increases beyond 20, these metrics are converged, indicating that \Tech{} can identify the best gradient within 20 trials and that additional trials yield plain enhancement.
Therefore, we set the default number of trials to 20.
Furthermore, in Figure~\ref{fig:ablation_trials_example}, while the attacker can invert some meaningful images with 1 and 5 trials, they fail to do so with 10 and 20 trials, as evidenced by the resulting noise in the inverted images.

\begin{figure}[ht!]
    \centering
    \includegraphics[width=.47\textwidth]{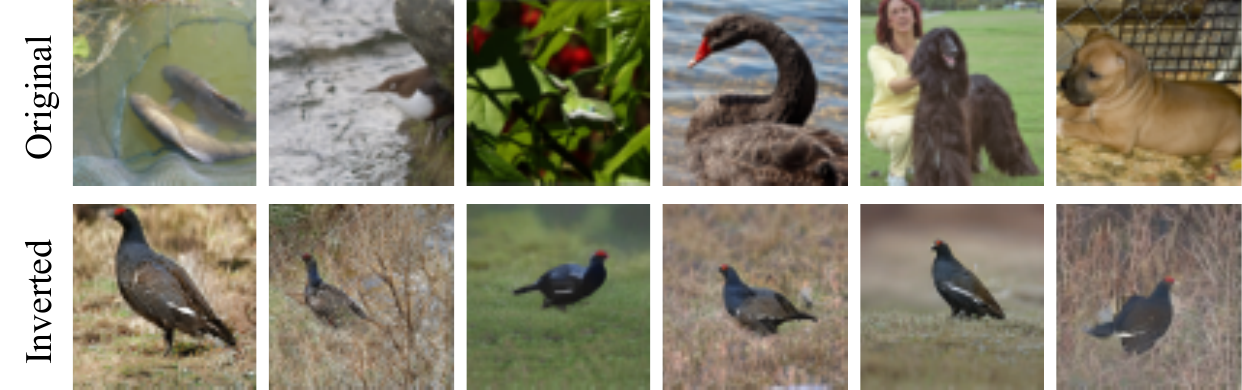}
    \caption{Gradient inversion under \Tech{}, with random gradient vectors constructed from ``Black grouse'' images.}
    \label{fig:other_instan_rand}
\end{figure}

\smallskip \noindent
\textbf{Effect of Sampling from a Public Dataset Other than a Normal Distribution $\mathcal{N}$.}
We examine the effect of sampling gradients from a publicly available dataset other than from the normal distributions $\mathcal{N}$ for \Tech{}. 
Typically, we utilize a normal distribution for $\mathcal{N}$, which demonstrates robust performance across various datasets and attacks, as shown in Table~\ref{tab:bigtable}.
In this experiment, we explore an alternative way to sample the perturbation that uses gradients derived from real images and evaluate the performance of \Tech{}.
The experiment is conducted on ImageNet using the GGL attack. Results, as presented in Figure~\ref{fig:other_instan_rand}, are based on constructing random gradient vectors from ``Black grouse'' images. 
We observe that the GGL~\cite{ggl} attack, which relies on GANs, is misled by our obfuscated gradients, resulting in the inversion of images that resemble black grouse, significantly different from the original training images.
This demonstrates that \Tech{}'s flexibility of accessing orthogonal gradient in different ways, as explained in theory part of Section~\ref{sec:method}, and is able to mislead the attackers in their inversion attempts.

\section{Related Work}\label{sec:related_work}

\smallskip \noindent
\textbf{Subspace Learning.}
Several works have investigated the geometry of the loss landscape of neural networks~\cite{tom_visualiz_loss,loss_surfaces}. \textit{Subspace learning} considers how neural networks can be optimized by exploring lower-dimensional subspaces within their high-dimensional parameter spaces.
These subspaces are defined as the set of parameter configurations near the current parameters that still achieve similar performance levels.
In particular, recent work shows there is significant freedom in the optimization paths that can achieve high accuracy models in large neural network models~\cite{orth_subspace, subspace_learning, goodfellow2014qualitatively,izmailov2020subspace}.
Wortsman et al.~\cite{subspace_learning} have notably identified and traversed large, diverse regions of the objective landscape through strategic exploration of these subspaces.

\smallskip \noindent
\textbf{Data Privacy.}
In federated learning,
recent studies indicate that 
user privacy can still be compromised in the presence of malicious attackers.
Existing works on membership inference attacks~\cite{mem_inference, shokri2017membership, carlini2022membership, salem2018ml, li2021membership, memberprivacy} involve a malicious entity determining whether a particular data sample was included in the training dataset. 
Subsequent studies have 
shown that the possibility of property inference attacks~\cite{property_inference_ccs,feature_leakage}.
It is furthermore explored on model inversion~\cite{model_inversion, zhang2020secret, yang2019neural, yin2020dreaming},
which elucidates the confidence scores output 
facilitates model inversion attacks.
Researchers find that recovering a recognizable face image of a person from shallow neural networks, using only their names and the model's output confidences~\cite{model_inversion}.
Further advancements in this area have demonstrated the feasibility of physical attacks~\cite{an2023imu,eykholt2018robust,an2024rethinking}.

\smallskip \noindent
\textbf{Defenses in Data Privacy.}
Various defense techniques have been proposed to counter privacy attacks in federated learning. Strategies such as reducing the overfitting of the global model~\cite{shokri2017membership,salem2018ml}, implementing differential privacy~\cite{yeom2018privacy,hui2021practical}, and masking confidence scores~\cite{shokri2017membership, jia2019memguard} have been 
proven effective in mitigating the risk of membership inference. 
To counter reconstruction attacks, 
researchers~\cite{bonawitz2017practical, lia2020privacy} employ multi-party computation to secure model updates.
Aggregation mechanisms~\cite{guo2021siren,zhang2023flip,guo2024siren+} are applied to defend the Byzantine attacks.
Existing studies~\cite{aono2017privacy, kim2018efficient} utilize Homomorphic Encryption to perform operations within the ciphertext space during gradient aggregation.

\section{Conclusion}\label{sec:conclusion}
We propose a novel defense technique, \Tech{}, designed to mitigate gradient inversion attacks. Our method samples gradients within a subspace orthogonal to the original gradients. By leveraging cold posteriors selection, \Tech{} employs a refined gradient update mechanism to enhance the data protection, while maintaining the model utility.
Our experiments show that \Tech{} substantially outperforms the state-of-the-art defenses, especially against advanced GAN-based attacks.

\section*{Acknowledgment}
We thank the anonymous reviewers for their constructive comments. We are grateful to the Center for AI Safety for providing computational resources. This work was funded in part by the National Science Foundation (NSF) Awards IIS-2229876, CNS-2247794, CAREER IIS-1943364, CNS-2212160, SHF-1901242, SHF-1910300, IIS-2416835, IARPA TrojAI W911NF-19-S0012, ONR N000141712045, N000141410468 and N000141712947. Any opinions, findings and conclusions or recommendations expressed in this material are those of the authors and do not necessarily reflect the views of the sponsors.

\bibliographystyle{IEEEtran}
\bibliography{reference}

\newpage
\appendix
\section{Appendix}\label{sec:appendix}

\subsection{Summary of Symbols}\label{appen:notations}
We summarize a comprehensive list of all notations in Table~\ref{tab:notation} for easy reference.
\begin{table}[ht]
\centering
\caption {Glossary of Notations.}
\label{tab:notation}
\begin{tabular}{lc}\toprule[1.5pt]
\textbf{Notation}                                                & \textbf{Description}                                                                                   \\\midrule
$f$                             & Model \\
$\theta$    & Model parameter\\
$\hat{\theta}$  & optimal model parameter\\
$\ell$                          & Loss function \\
$\mathbb{R}^d$                  & Input space   \\
$\mathbb{R}^k$                  & Latent space of the generative model   \\
$N$         & Number of local clients \\
$k$         & $k$-th local client \\
$n$         & Batch size on the client\\
$\tau$      & Number of federated learning iteration \\
$X', Y'$    & Randomly initialized inputs \\
$\hat{X'}, \hat{Y'}$  & Reconstructed inputs \\
$\mathcal{D}(\cdot)$   & Distance metric \\
$G_p(\cdot)$           & Pre-trained generative model   \\
$\mathcal{T}(\cdot)$          & Gradient transformation function    \\
$\phi(\cdot)$  & Regularization term \\
$D$         & Dataset \\
$P(\cdot | \cdot)$         & Normalized probabilities \\
$M$         & Temperature paramete \\
$T$         & Sampling trials \\
$\mathcal{N}({\bf 0}, \cdot)$                   & Normal distribution \\
$\perp$     & Orthogonal \\
$\langle \cdot,\cdot\rangle$ & Inner product\\
$G^0$       & Original gradient\\
$G^{\star}$  & Best gradient    \\
$z$         & Latent space of the generative model \\
$C$         & Number of classes \\
$W_{\text{FC}}$ & Weights of final fully-connected (FC) classification layer\\
$g^o$                             & Orthogonal gradients \\
$g_r$                             & Random gradients \\
$l$                             & Layer \\

\bottomrule[1.25pt]
\end{tabular}\par
\end{table}

\ndssrevise{
\subsection{Overhead Evaluation}\label{appen:overhead}
In this section, we conduct an evaluation to quantify the computational overhead and additional processing latency of \Tech{} and Soteria~\cite{soteria}. 
We define the overhead percentage as:
Overhead Percentage = ( (Time with Defense - Time without Defense) / Time without Defense)$\times$100\%.
Applying this formula to \Tech{}, the overhead calculation yields an exceptionally low impact of approximately 0.00236\%. Comparatively, we also assessed Soteria, which introduces an overhead of approximately 0.3092\%, significantly higher than \Tech{}. 
}

\begin{table}[ht]
\centering
\tabcolsep=7pt
\caption{Ablation study for different number of rounds.}
\label{tab:ablation_ours_effective}
\begin{tabular}{ccccc}
\toprule
\multirow{2}{*}{\textbf{Round ID}} & \multicolumn{4}{c}{\textbf{GIFD}}                                          \\
\cmidrule(lr){2-5}
                                 & \textbf{MSE ↑} & \textbf{LPIPS ↑} & \textbf{PSNR ↓} & \textbf{SSIM ↓} \\
\midrule
0                                & 0.0507         & 0.7610                 & 13.32           & 0.009          \\
1                                & 0.0989         & 0.7368                 & 10.50           & 0.036          \\
2                                & 0.1224         & 0.7394                 & 9.94            & 0.036          \\
3                                & 0.1040         & 0.7370                 & 10.37           & 0.042          \\
4                                & 0.1717         & 0.7528                 & 9.07            & 0.041          \\
5                                & 0.1233         & 0.7296                 & 10.16           & 0.035          \\
6                                & 0.1049         & 0.7272                 & 10.28           & 0.034          \\
7                                & 0.1232         & 0.7425                 & 9.95            & 0.033          \\
8                                & 0.1067         & 0.7173                 & 10.69           & 0.041          \\
9                                & 0.1034         & 0.7247                 & 10.65           & 0.038         \\

\bottomrule
\end{tabular}
\end{table}

\subsection{Performance of \Tech{} Applied for Different Number of Training Rounds.} \label{appen:addi_1}
We investigate the impact of applying \Tech{} across various numbers of training rounds to assess its performance under different conditions. This experiment is conducted using the ImageNet dataset and the GIFD attack. We systematically apply \Tech{} from 0 to 9 training rounds and measure the performance of the attack at each interval. The results of this study are detailed in Table~\ref{tab:ablation_ours_effective}, where we present four quantitative metrics for each training round to provide a comprehensive evaluation.
According to the findings presented in Table~\ref{tab:ablation_ours_effective}, \Tech{} consistently demonstrates effectiveness in countering the GIFD attack through various stages of training. Notably, as the number of training rounds increases from 0 to 9, we observe a noticeable increase in the MSE loss from 0.0507 to 0.1034, while the PSNR decreases from 13.32 to 10.65, it still outperforms existing defenses. 
This improvement in defensive capability against the GIFD attack not only highlights the robustness of \Tech{} but also corroborates our earlier observation documented in Section~\ref{sec:observations}. Specifically, as the training progresses and the model approaches convergence, it becomes progressively more difficult for an attacker to successfully reconstruct raw data from the gradients.

\ndssrevise{
\subsection{Performance of \Tech{} Across Different Number of Clients.} \label{appen:num_of_clients}
We evaluate the performance of \Tech{} in different numbers of participating clients. Specifically, our experimental setup involves 100 clients by default.
and we test our defense with 10, 30, and 50 selected clients out of 100 on the CIFAR-10 dataset under a \textit{non-i.i.d.} setting. 
Results in Table~\ref{tab:appen_number_of_clients} show that our technique consistently converges and matches the performance of the vanilla aggregated model without defense, regardless of the number of clients. 
This suggests that our defense scales effectively with the number of clients and does not necessarily lead to a substantial decrease in performance.

\begin{table}[ht]
\centering
\tabcolsep=7pt
\caption{\ndssrevise{Performance of \Tech{} across different number of clients.}}
\label{tab:appen_number_of_clients}
\begin{tabular}{cc}
\toprule
\multicolumn{1}{l}{Number of Clients} & \multicolumn{1}{l}{Accuracy} \\
\midrule
10                                    & 85.27                        \\
30                                    & 85.33                        \\
50                                    & 85.30                       \\
\bottomrule
\end{tabular}
\end{table}

}

\ndssrevise{
\subsection{Discussion of Adaptive Attacks}\label{appen:adaptive}
In this section, we discuss additional possible adaptive attacks. 
First, leveraging a similar mechanism as \Tech{} to train a Generative Adversarial Networks (GAN) and recover the original gradients. Existing GAN models~\cite{ffhq, biggan} utilize considerably smaller latent spaces, although it is non-trivial to employ the gradient as a latent space to train a GAN to reconstruct the original gradients effectively in the context of \Tech{}, it can be explored in the future. 
Second, given an overparameterized model, there exists many different directions that can improve the loss. If the attacker has knowledge about the cold posterior sampling mechanism, they might attempt to manipulate the temperature parameter. By injecting specially crafted updates that influence the posterior distribution, attackers could bias the sampling process, making it easier to infer the original gradients.
}

\subsection{Ablation Study}~\label{appen:abalation}

\begin{figure}[ht]
    \centering
    \includegraphics[width=.35\textwidth]{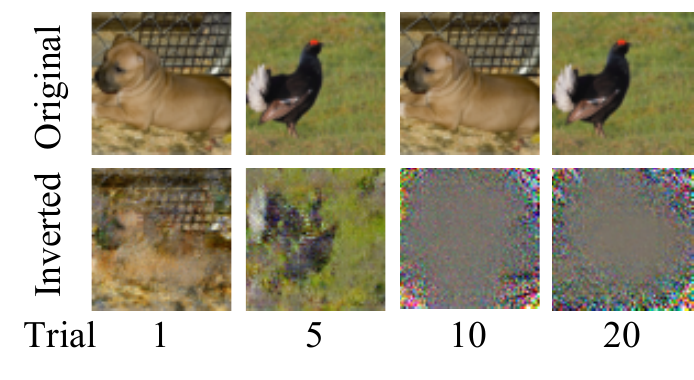}
    \caption{Trials examples on GIAS.}
    \label{fig:ablation_trials_example}
\end{figure}

\noindent
\textbf{Effect of Applying Layer-wise Operation.}
We explore the impact of applying the layer-wise operation of \Tech{}. The experiment, conducted on ImageNet using the GIFD attack, is detailed in Table~\ref{tab:ablation_layerwise}. We assess the inversion performance of the attack using four metrics. Results indicate that the layer-wise application of \Tech{} slightly outperforms the technique when applied to the entire gradient vector of the whole model.
This improvement can be attributed to the fine-grained orthogonal projection and normalization processes detailed in Section~\ref{sec:method}, which enhance the specificity and efficacy of the defense by adapting it more precisely to the unique characteristics of each layer.

\begin{table}[ht]
    \centering
    \tabcolsep=7pt
    \caption{Ablation study on layer-wise operation.}
    \label{tab:ablation_layerwise}
    \begin{tabular}{lcccc}
        \toprule
        \textbf{Config}  & \multicolumn{1}{c}{\textbf{MSE ↑}} & \multicolumn{1}{c}{\textbf{LPIPS ↑}} & \multicolumn{1}{c}{\textbf{PSNR ↓}} & \multicolumn{1}{c}{\textbf{SSIM ↓}} \\
        \midrule
        Layer-wise gradient     & \textbf{0.0507}                    & \textbf{0.7610}                            & \textbf{13.32}                      & \textbf{0.009}                     \\
        Entire gradient & 0.0452                    & 0.7532                            & 13.81                      & 0.012                    \\
        \bottomrule
    \end{tabular}
\end{table}

\end{document}